%% file: top.tex
\documentclass{article}

\usepackage{microtype}
\usepackage{graphicx}
\usepackage{subcaption}
\usepackage{booktabs} 

\usepackage{hyperref}

\usepackage[accepted]{icml2026}

\usepackage{amsmath}
\usepackage{amssymb}
\usepackage{mathtools}
\usepackage{amsthm}

\usepackage[capitalize,noabbrev]{cleveref}

\theoremstyle{plain}

\theoremstyle{definition}

\theoremstyle{remark}

\usepackage[textsize=tiny]{todonotes}
\usepackage{listings}
\usepackage{wrapfig}
\usepackage{multirow}
\icmltitlerunning{UniRTL: Unifying Code and Graph for Robust RTL Representation Learning}

\begin{document}

\twocolumn[
  \icmltitle{UniRTL: Unifying Code and Graph for Robust RTL Representation Learning}



  \icmlsetsymbol{equal}{*}

  \begin{icmlauthorlist}
    \icmlauthor{Yi Liu}{equal,cuhk}
    \icmlauthor{Hongji Zhang}{equal,cuhk}
    \icmlauthor{Lei Chen}{huawei}
    \icmlauthor{Mingxuan Yuan}{huawei}
    \icmlauthor{Qiang Xu}{cuhk}
  \end{icmlauthorlist}

  \icmlaffiliation{cuhk}{Department of Computer Science and Engineering, The Chinese University of Hong Kong, Hong Kong SAR}
  \icmlaffiliation{huawei}{Noah's Ark Lab, Huawei, Hong Kong SAR}

  \icmlcorrespondingauthor{Qiang Xu}{qxu@cse.cuhk.edu.hk}

  \icmlkeywords{ICML, Multimodal Representation Learning}

  \vskip 0.3in
]



\printAffiliationsAndNotice{\icmlEqualContribution}

\input{secs/0_abstract}
\input{secs/1_introduction}
\input{secs/2_related_work}
\input{secs/3_methodology}
\input{secs/4_experiment}
\input{secs/5_limitations}
\input{secs/6_conclusion}

\section*{Acknowledgments}
This work was supported in part by the Hong Kong Research Grants Council (RGC) under Grant No. 14202824, C6003-24Y, and T46-415/25-R, and in part by Huawei under Grant No. N2-2c-TH2420350.

\section*{Impact Statement}
This paper presents work whose goal is to advance the field of Machine
Learning. There are many potential societal consequences of our work, none
which we feel must be specifically highlighted here.

\bibliography{top}
\bibliographystyle{icml2026}

\newpage
\appendix
\onecolumn
\section{Analysis of CDFG Conversion Failures and Subset Representativeness}
\label{appendix:cdfg_conversion_failure_analysis}
In Section~\ref{subsec:dataset_construction}, we note that not all collected RTL designs can be successfully converted into CDFGs. Many designs originate from open-source GitHub repositories or are generated by LLMs and contain syntax errors that lead to compilation failures. Typical failure cases include malformed or incomplete RTL modules, invalid declarations, inconsistent interfaces, and constructs that prevent successful parsing or compilation by the Yosys-based conversion pipeline. In total, our dataset contains 132,008 RTL designs, of which 38,888 are successfully converted into CDFGs.

Importantly, the 38,888 designs that successfully convert to CDFGs do not constitute a cherry-picked subset of simpler or cleaner RTL. In practice, most conversion failures arise from low-quality designs in existing open-source corpora, which would also fail standard EDA toolchains. Consequently, our filtering primarily enforces basic syntactic and compilation validity rather than favoring structurally simple circuits, and the resulting corpus still spans a broad range of design sizes, structural complexity, and coding styles.

\begin{figure}[t]
    \centering
    \includegraphics[width=0.9\linewidth]{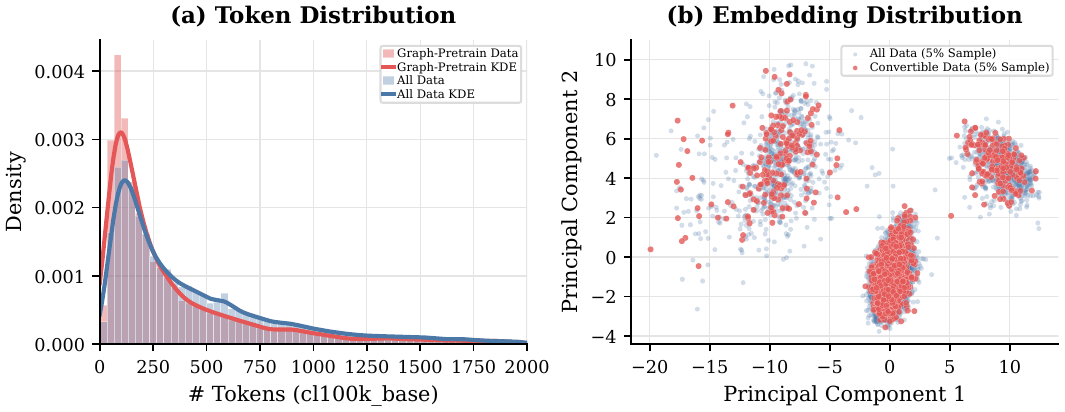}
    \caption{Representativeness analysis of the CDFG-convertible subset. We compare the full pretraining dataset and the subset of designs successfully converted into CDFGs using token-count distributions and UniRTL embedding distributions.}
    \label{fig:cdfg_convertible_representativeness}
\end{figure}

To further examine the representativeness of the CDFG-convertible subset, we compare it with the full pretraining dataset from both scale and representation perspectives. For scale, we use Gaussian kernel density estimation (KDE) to compare token-count distributions, where tokens are obtained with the \texttt{cl100k\_base} tokenizer from OpenAI's tiktoken library\footnote{\url{https://github.com/openai/tiktoken}}. For representation, we randomly sample 5\% of examples from both the full dataset and the CDFG-convertible subset, extract embeddings using the pretrained UniRTL model, and apply PCA~\citep{mackiewicz1993principal} for two-dimensional visualization. As shown in Figure~\ref{fig:cdfg_convertible_representativeness}, the token-count distributions are highly similar, and the two groups occupy largely overlapping regions in the representation space. This suggests that the CDFG-convertible subset is broadly representative of the full pretraining dataset rather than being biased toward only simple designs.

Nevertheless, we retain the noisy samples that fail CDFG conversion for text-code alignment, enabling the model to learn more robust and generalizable representations while maximizing data utilization. Empirical validation of this design choice is provided in Appendix~\ref{appendix:leveraging_noisy_samples_for_text_code_alignment}.

\section{Details of the Graph-Aware Tokenizer}
\label{appendix:graph_aware_tokenizer_details}
The graph-aware tokenizer integrates a graph isomorphism network (GIN)~\citep{xupowerful} with a lightweight Transformer encoder to jointly capture local structural dependencies and global contextual information.
It is pretrained with two objectives, structure-aware masked node modeling and edge prediction, which enable it to capture the nuanced and intricate structural relationships within the graph.
Specifically, given the initial node embeddings $\{\mathbf{H}_{i}\}_{i\in\mathbb{V}}$, the graph is first processed by the GIN to obtain $\{\mathbf{L}_{i}\}_{i\in\mathbb{V}}$ that encode local structural dependencies. These embeddings are then passed through the Transformer encoder to produce refined node embeddings $\{\mathbf{G}_{i}\}_{i\in\mathbb{V}}$.
For structure-aware masked node modeling, we randomly replace 20\% of nodes with a special learnable \texttt{[MASK]} embedding at the post-GIN level and use the Transformer encoder to recover masked nodes by predicting their original node types.
Following StructRTL~\citep{liu2025beyond}, we adopt the class-balanced focal loss~\cite{cui2019class} for this task to mitigate the node-type imbalance problem and denote the loss as $\mathcal{L}_{mnm}$.
For edge prediction, the refined node embeddings $\{\mathbf{G}_{i}\}_{i\in\mathbb{V}}$ are used to predict the existence of edges between nodes.
Since the Transformer encoder discards explicit connectivity, which can be viewed as if all edges are masked, we sample 20\% of true edges as positive samples and an equal number of non-existing edges as negative samples in each iteration. The task is formulated as a binary classification problem, where we concatenate the final embeddings of the source and target nodes and use a three-layer multi-layer perceptron (MLP) to predict whether an edge exists between them.
The cross-entropy loss is employed for this task, and the loss is denoted as $\mathcal{L}_{ep}$.
Overall, the graph-aware tokenizer is pretrained with the loss:
\begin{equation}
    \mathcal{L} = \gamma \cdot \mathcal{L}_{mnm} + (1 - \gamma) \cdot \mathcal{L}_{ep}
\end{equation}
where $\gamma$ balances these two pretraining tasks, with $\gamma = 0.5$ in our experiments.

When node embeddings are flattened for input into the Transformer encoder, the graph's topological information is lost. 
To mitigate this issue, we incorporate global positional encodings into the post-GIN node embeddings $\{\mathbf{L}_{i}\}_{i\in\mathbb{V}}$ before feeding them into the Transformer encoder. The construction and application of these global positional encodings are described in Section~\ref{sec:model_architecture}.

The graph-aware tokenizer employs an 8-layer GIN and an 8-layer Transformer encoder, with 4 attention heads per Transformer layer.
It is pretrained for 2,000 epochs with a batch size of 16 on a single NVIDIA L40 GPU, using AdamW~\citep{loshchilovdecoupled} with a learning rate of 2e-5 and weight decay of 1e-4.

After pretraining, the graph-aware tokenizer achieves evaluation accuracies of 85.04\% on the structure-aware masked node modeling task and 99.68\% on the edge prediction task.
Following UniRTL pretraining, the masked node recovery accuracy further improves to 97.57\%, demonstrating that incorporating code information enhances recovery performance and validates the effectiveness of our alignment strategy.
We do not incorporate the edge prediction task during the pretraining of UniRTL since this task is relatively simple, converge quickly to high accuracy, and has negligible impact on the final model performance.

\section{Efficiency Comparison}
\label{appendix:efficiency_comparison}

We provide a detailed efficiency comparison for UniRTL and the baselines in Table~\ref{tab:training_time_comparison}, explicitly separating costs from pretraining, downstream fine-tuning, and inference. The pretraining columns report the cost of representation-model pretraining when applicable, while the fine-tuning columns report downstream task adaptation costs. Entries marked ``None'' indicate that the corresponding stage is not used by that method, and ``N/A'' indicates that the cost is not available, such as for API-only models.

Compared with LLM-based encoders such as GritLM-7B and DeepRTL2, UniRTL requires substantially less pretraining compute: it is pretrained on 2$\times$ NVIDIA L40 GPUs for approximately 45 hours, whereas DeepRTL2 uses 8$\times$ NVIDIA A800 GPUs for 70 hours and GritLM-7B uses 64$\times$ NVIDIA A100 GPUs for 48 hours. GraphCodeBERT also incurs a higher pretraining cost, requiring 16$\times$ NVIDIA V100 GPUs for 83 hours. At inference time, UniRTL requires about 300 MB of bfloat16 memory, which is much smaller than 7B-scale embedding or RTL-code models and only modestly larger than GraphCodeBERT and StructRTL. Overall, UniRTL attains strong performance with a moderate training budget and a compact inference footprint compared with large LLM-based encoders.

\input{tabs/training_time_comparison}

\section{Fine-Tuning for Performance Prediction}
\label{appendix:performance_prediction_details}
After obtaining RTL representations from different methods, we fine-tune a three-layer MLP for performance prediction. 
Because the dimensionality of RTL representations varies across methods, we first project them into a 512-dimensional space before feeding them into the MLP, which has a hidden layer size of 256.
Given that area and delay values have large magnitudes and exhibit substantial variance across designs, we follow VeriDistill~\citep{moravej2025graph} and StructRTL~\citep{liu2025beyond} to apply a logarithm transformation to these values, making the target distribution more suitable for model learning.
This transformation does not affect the practical utility of the predictor, as we are more concerned with the relative quality of different designs.

For training, we adopt the log-cosh loss~\citep{saleh2022statistical}, which is robust to outliers. The three-layer MLP predictors are trained for 600 epochs on a single NVIDIA L40 GPU with a batch size of 256, using the Adam optimizer~\citep{kingma2014adam} with a learning rate of 1e-4 and a weight decay of 1e-5. Under this setup, all models are trained until full convergence.

\section{Evaluation Metrics for Performance Prediction}
\label{appendix:evaluation_metrics_for_performance_prediction}
For performance prediction evaluation, we employ four standard regression metrics: mean absolute error (MAE), mean absolute percentage error (MAPE), coefficient of determination ($R^{2}$), and root relative squared error (RRSE). 
Given predicted values $\hat{y_{i}}$ and ground truth values $y_{i}$ for $i \in \left[1,N\right]$, these metrics are defined as:
\begin{align}
\text{MAE} &= \frac{1}{N} \sum_{i=1}^{N} \left| \hat{y}_i - y_i \right| \\
\text{MAPE} &= \frac{1}{N} \sum_{i=1}^{N} \left| \frac{\hat{y}_i - y_i}{y_i} \right| \\
R^2 &= 1 - \frac{\sum_{i=1}^{N} (\hat{y}_i - y_i)^2}{\sum_{i=1}^{N} (y_i - \bar{y})^2} \\
\text{RRSE} &= \sqrt{\frac{\sum_{i=1}^{N} (\hat{y}_i - y_i)^2}{\sum_{i=1}^{N} (y_i - \bar{y})^2}}
\end{align}
where $\bar{y}$ denotes the mean of the ground truth values.

\input{tabs/nlcs_hyperparameters}

\section{Performance Prediction with Netlist Information}
\label{appendix:performance_prediction_with_netlist_info}
To further enhance performance prediction, we incorporate a knowledge distillation strategy that transfers low-level insights from post-mapping (PM) netlists into the RTL-stage performance predictors, \emph{i.e.}, the three-layer MLP described in Appendix~\ref{appendix:performance_prediction_details}.
Following StructRTL~\citep{liu2025beyond}, we collect all synthesized PM netlists and train a GIN to directly predict performance metrics from these netlists.
A PM netlist typically consists of interconnected logic cells defined in a technology library.
To represent the PM netlist, we initialize each cell’s embedding as the concatenation of its one-hot cell type encoding, logic truth table, and associated area and pin delay information.
These embeddings are then processed by the GIN, followed by joint mean and max pooling to produce a graph-level representation, which is subsequently fed into a three-layer MLP for performance estimation.
After training the PM predictor, we freeze its parameters and introduce a knowledge distillation loss during the training of the RTL-stage predictor, aligning the final-layer activations of the RTL-stage predictor ($z_{\text{RTL}}^{-1}$) with those of the PM predictor ($z_{\text{PM}}^{-1}$). The knowledge distillation loss is defined as:
\begin{equation}
\mathcal{L}_{kd} = \alpha \cdot \mathcal{L}_{cos}(z_{\text{RTL}}^{-1}, z_{\text{PM}}^{-1}) + (1 - \alpha) \cdot \mathcal{L}_{mse}(z{\text{RTL}}^{-1}, z_{\text{PM}}^{-1})
\end{equation}
where $\mathcal{L}_{cos}$ denotes the cosine similarity loss, $\mathcal{L}_{mse}$ the mean squared error (MSE) loss, and $\alpha$ balances the contribution of these two loss terms, set to 0.7 in our experiments.

The final loss for the RTL-stage predictor combines this distillation term with the log-cosh loss described in Appendix~\ref{appendix:performance_prediction_details}:
\begin{equation}
    \mathcal{L}_{pred} = \beta \cdot \mathcal{L}_{log\_cosh} + (1 - \beta) \cdot \mathcal{L}_{kd}
\end{equation}
where $\beta$ is set to 0.5 in our experiments.

The adopted GIN consists of 20 layers with residual connections and is trained for 1,000 epochs using the log-cosh loss, a batch size of 16, and the same optimizer configuration as the RTL-stage predictor, on a single NVIDIA L40 GPU.
It is important to note that the PM predictor is only used during training as a teacher; during inference, only the RTL-stage performance predictor is retained.

\input{tabs/few_shot_performance_prediction}

\input{tabs/incomplete_pretrain_performance_prediction}

\section{Few-Shot Fine-Tuning for Performance Prediction}
\label{appendix:few_shot_finetuning_performance_prediction}
Performance labels are tied to the synthesis setup used to generate them, including the process design kit and standard cell library. Therefore, a predictor trained under one setup may not remain fully accurate in a zero-shot manner under a substantially different setup, such as when migrating to a new technology node. This challenge applies to performance prediction generally.

Although we do not conduct a direct cross-node transfer experiment in this work, UniRTL is designed to reduce the cost of migration across synthesis setups. The UniRTL backbone is pretrained on RTL code, functional summaries, and CDFGs, and therefore does not depend on a particular process node. If labels from a new synthesis flow become available, the backbone does not need to be pretrained again; instead, one can reuse it and fine-tune only the lightweight performance prediction head. If netlist distillation is used, the PM-netlist teacher and distillation stage can similarly be updated as downstream components rather than requiring full retraining from scratch.

As supporting evidence that the pretrained representation can be adapted with limited labeled data, we conduct a few-shot performance prediction study using different proportions of the original training set without netlist information. Table~\ref{tab:few_shot_performance_prediction} shows that UniRTL improves consistently as more labeled data becomes available, and already achieves competitive performance with only 20\% of the training data. This experiment does not fully resolve transfer across synthesis setups, but it indicates that the cost of obtaining a useful predictor under a new setup can be mitigated by reusing the pretrained backbone and performing lightweight fine-tuning with limited newly synthesized data.

\input{tabs/incomplete_pretrain_natural_language_code_search}

\input{tabs/incomplete_pretrain_functionality_equivalence_checking}

\section{Experimental Setup for Natural Language Code Search}
\label{appendix:nlcs_experimental_setup}
We employ different strategies to obtain embeddings for functional summaries and RTL designs when evaluating different models on the natural language code search task. Specifically, for general-purpose text embedding models and customized RTL embedding models, we directly use their pretrained weights and provided APIs to generate embeddings of the functional summaries and RTL designs. Since GritLM-7B~\citep{muennighoff2025generative} and NV-Embed-v2~\citep{lee2025nv} are trained under an instruction-tuning paradigm, we prepend the instruction ``\textit{Given a high-level functional summary, retrieve the corresponding RTL code.}'' in the model-specific template when extracting embeddings of functional summaries.

For GraphCodeBERT~\citep{guo2021graphcodebert} and UniRTL, we take the last hidden state of the first token, \textit{i.e.}, the \texttt{[CLS]} token, as the embedding vector for both $\mathcal{S}_i$ and $\mathcal{R}_i$ (in either code-only or ``code \& graph" format). These models are fine-tuned on this task using contrastive learning prior to evaluation. For all model variants, we keep the dataset and hyperparameter settings consistent during fine-tuning to ensure a fair comparison.
We adopt the InfoNCE loss~\citep{oord2018representation} for downstream fine-tuning on this task:
\begin{equation}
\mathcal{L}_{\text{nlcs}}=-\frac{1}{M}\sum_{i=1}^{M}\log\left(\frac{\exp\left(\frac{\operatorname{cos}(f_{\theta}(\mathcal{S}_i), f_{\theta}(\mathcal{R}_i))}{\tau}\right)}{\sum_{j=1}^{M}\exp\left(\frac{\operatorname{cos}(f_{\theta}(\mathcal{S}_i), f_{\theta}(\mathcal{R}_j))}{\tau}\right)}\right)
\end{equation}
where $M$ is the batch size, $\mathcal{S}_i$ is the $i$-th functional summary in the batch, $\mathcal{R}i$ is the corresponding RTL design, $f_\theta$ is the embedding function, and $\tau$ is the temperature hyperparameter.

Let the evaluation benchmark be $(\mathcal{S}, \mathcal{R})$, where both $\mathcal{S}$ and $\mathcal{R}$ contain $N$ samples. During evaluation, the task is formulated as an $N$-class classification problem. Each $\mathcal{S}_i$ is treated as a sample belonging to class $i$, and the embedding model $f_\theta$ predicts its class as $\underset{k}{\arg\max}\operatorname{cos}(f_\theta(\mathcal{S}_i), f_\theta(\mathcal{R}_k))$. 

Evaluation metrics for this task include Precision, Recall and F1, following the standard paradigm of multi-class classification, with F1 serving as the main metric.
Downstream fine-tuning for this task is conducted on a single NVIDIA L40 GPU, and the hyperparameter settings are provided in Table~\ref{tab:nlcs_hyperparameters}. An illustrative data example for this task is shown in Listing~\ref{lst:nlcs_data_example}, comprising a high-level functional summary of an arithmetic logic unit (ALU) and its corresponding Verilog implementation.

\input{tabs/alignment_ablation_performance_prediction}

\input{tabs/alignment_ablation_natural_language_code_search}

\section{Experimental Setup for Functionality Equivalence Checking}
\label{appendix:fec_experimental_setup}
For the functionality equivalence checking task, we follow a strategy similar to that used for natural language code search to obtain embeddings of RTL designs.
General-purpose text embedding models and customized RTL embedding models are evaluated without tuning their original parameters. For instruction-tuned text embedding models GritLM-7B~\citep{muennighoff2025generative} and NV-Embed-v2~\citep{lee2025nv}, we prepend the instruction ``\textit{Determine whether the given pair of RTL code snippets is functionally equivalent.}'' to their model-specific templates to adapt their embeddings to this task.

For GraphCodeBERT~\citep{guo2021graphcodebert} and UniRTL, we take the last hidden state of the \texttt{[CLS]} token as the embedding vector for each RTL design. These models are fine-tuned on this task using contrastive learning, where functionally inequivalent designs are used as hard negatives. To ensure fair comparison, all variants are fine-tuned under identical dataset and hyperparameter settings.

The fine-tuning dataset for this task is formatted as $\{(\mathcal{R}_i, \mathcal{E}_i, \mathcal{U}_i)\}_{i=1}^{N}$, where $\mathcal{R}_i$ is an RTL design, $\mathcal{E}_i$ is a corresponding RTL design with the same functionality, and $\mathcal{U}_i$ is a list of functionally inequivalent designs that serve as hard negatives.
We adopt the  InfoNCE loss~\cite{oord2018representation} with hard negatives for downstream fine-tuning on this task:
\begin{equation}
\mathcal{L}_{\text{fec}}=-\frac{1}{M}\sum_{i=1}^{M}\log\left(\frac{\exp\left(\frac{\operatorname{cos}(f_\theta(\mathcal{R}_i), f_\theta(\mathcal{E}_i))}{\tau}\right)}{\sum_{j=1}^{M}\exp\left(\frac{\operatorname{cos}(f_\theta(\mathcal{R}_i), f_\theta(\mathcal{E}_j)))}{\tau}\right)+\sum_{j=1}^{M}\sum_{k=1}^{h_j}\left(\frac{\operatorname{cos}(f_\theta(\mathcal{R}_i), f_\theta(\mathcal{U}_j[k]))}{\tau}\right)}\right)
\end{equation}
where $M$ is the batch size, $f_\theta$ is the embedding function, $\tau$ is the temperature hyperparameter, and $h_j = \min(\operatorname{length}(\mathcal{U}_j), \text{max\_hard\_negatives})$, is the number of hard negatives used for sample $j$, controlled by the hyperparameter max\_hard\_negatives.

We evaluate models using five metrics: Average Precision (AP), Accuracy, F1, Precision, and Recall, with AP serving as the main metric. All evaluation metrics take as input a list of cosine similarity scores and binary labels, where 1 indicates functional equivalence and 0 indicates inequivalence. 
The threshold for functional equivalence is determined differently depending on the specific metric. The main metric, AP, requires no thresholding and is computed using the \texttt{average\_precision\_score} function in the Python's \texttt{scikit-learn} library\footnote{\url{https://scikit-learn.org/stable/modules/generated/sklearn.metrics.average_precision_score.html}}.
For accuracy, the threshold that maximizes classification accuracy is selected by enumerating over all possible thresholds. 
Specifically, we rank the similarity scores from highest to lowest, compute the accuracy at each possible threshold, and select the threshold that achieves the maximum accuracy. For F1, we similarly enumerate thresholds to identify the one that maximizes F1, and then report the corresponding F1, Precision, and Recall. This process ensures that we use the most appropriate threshold for each metric, allowing for accurate evaluation for this task. Our evaluation pipeline follows the pair-classification paradigm of the MTEB benchmark~\citep{muennighoff2023mteb}, with implementation details available in the official MTEB GitHub repository\footnote{\url{https://github.com/embeddings-benchmark/mteb}}.
Downstream fine-tuning for this task is performed on a single NVIDIA L40 GPU, and the hyperparameter settings are listed in Table~\ref{tab:fec_hyperparameters}.
An example training instance from the functionality equivalence checking dataset is shown in Listing~\ref{lst:fec_data_example}. In this example, all three RTL designs share the same module name and interface (inputs and outputs), but differ in their internal implementations. The ``Code’’ design serves as the query, the ``Equal’’ design has equivalent functionality, and the ``Unequal’’ design is functionally inequivalent to the query design despite structural similarity.

\input{tabs/alignment_ablation_functionality_equivalence_checking}

\section{Leveraging Noisy Samples for Text-Code Alignment}
\label{appendix:leveraging_noisy_samples_for_text_code_alignment}
In Appendix~\ref{appendix:cdfg_conversion_failure_analysis}, we state that RTL samples that cannot be successfully converted into CDFGs are still used as noisy data for text-code alignment, enabling the model to learn more robust and generalizable representations while maximizing data utilization. To assess whether this strategy truly improves learning and makes effective use of the available data, we compare UniRTL (w/o graph) with two CodeBERT-based baselines: the original CodeBERT-base-mlm model (CodeBERT) and CodeBERT-IP (Incomplete Pretrain), a variant obtained by pretraining and fine-tuning CodeBERT-base-mlm only on the 38,888 designs with valid CDFGs, while keeping all other settings fixed. As shown in Tables~\ref{tab:incomplete_pretrain_performance_prediction},~\ref{tab:incomplete_pretrain_natural_language_code_search} and~\ref{tab:incomplete_pretrain_functionality_equivalence_checking}, CodeBERT-IP consistently outperforms the vanilla CodeBERT baseline, demonstrating the effectiveness of our curated RTL dataset constructed from designs with valid CDFGs. Moreover, UniRTL (w/o graph), which further leverages noisy samples without CDFGs, surpasses both CodeBERT and CodeBERT-IP. This indicates that our hierarchical training strategy genuinely maximizes data utilization and yields better representations, rather than merely masking a data-quality problem.

\section{Ablation Study of Code–Graph Alignment Strategies}
\label{appendix:ablation_study_of_code_graph_alignment_strategies}
In addition to the modality ablations reported in the main paper, we further analyze how different code–graph alignment strategies affect UniRTL’s performance. First, we note that GraphCodeBERT is already included as a strong baseline that reflects an alternative alignment mechanism between code and graph. To ensure a fair comparison, we fine-tune GraphCodeBERT on our dataset rather than using its original, software-oriented form. GraphCodeBERT aligns code and graph via a contrastive variable-alignment objective that encourages correspondence between variable nodes in the graph and code tokens. While effective, this objective primarily targets variable-level matching and does not explicitly model richer semantic relations between arbitrary tokens and CDFG nodes.
By contrast, UniRTL adopts a mutual masked modeling objective across the code and CDFG modalities. Rather than aligning only variable mentions, UniRTL jointly reconstructs masked elements in one modality using information from the other, thereby encouraging finer-grained cross-modal alignment between code tokens and graph nodes. This design is intended to capture broader semantic and structural relationships beyond simple variable correspondences.

To more directly probe the effect of this mutual masked modeling objective, we conduct an additional ablation with a “direct-combine” baseline. In this variant, we simply concatenate the representations from the text–code-aligned encoder and the graph-aware tokenizer, without any explicit multimodal alignment pretraining between code and graph. All models are trained and evaluated under identical settings.
As shown in Tables~\ref{tab:alignment_ablation_performance_prediction},~\ref{tab:alignment_ablation_natural_language_code_search}, and~\ref{tab:alignment_ablation_functionality_equivalence_checking}, across all tasks, the direct-combine baseline improves over the corresponding unimodal variant, confirming that the mere presence of both modalities is beneficial. However, UniRTL with mutual masked modeling consistently achieves the best performance, providing clear and non-trivial gains over direct combine on natural language code search, functionality equivalence checking, and both area and delay prediction. 
These improvements indicate that our alignment strategy contributes beyond mere multi-modality by enabling finer-grained cross-modal understanding between code and CDFG.

\newpage

\begin{lstlisting}[basicstyle=\ttfamily\small, keywordstyle=\color{black}, stringstyle=\color{black}, commentstyle=\color{black}, numbers=none, showstringspaces=false, tabsize=4, breaklines=true, breakindent=0pt, frame=single, caption={Functional summary of an ALU and its corresponding Verilog implementation.}, label=lst:nlcs_data_example, columns=fullflexible, keepspaces=true, xleftmargin=0pt, xrightmargin=0pt, aboveskip=0pt, belowskip=0pt]
Functional Summary:
The code defines an ALU that executes operations like add, subtract, bitwise AND/OR, left/right shift, and bitwise NOT based on a control signal. It processes two 8-bit inputs and produces an 8-bit result, with a flag to indicate a zero output.
\end{lstlisting}

\begin{lstlisting}[language=Verilog, basicstyle=\ttfamily\small, keywordstyle=\color{blue}\bfseries, stringstyle=\color{purple}, commentstyle=\color{gray}\itshape, numbers=none, showstringspaces=false, tabsize=4, breaklines=true, frame=single, columns=fullflexible, keepspaces=true, xleftmargin=0pt, xrightmargin=0pt, aboveskip=0pt]
Code:
module Alu(
    Alu_in1,       
    Alu_in2,       
    Alu_sel,       
    Alu_zero_flg,  
    Alu_out        
);
    parameter wrd_size = 8,  
               sel_width= 3; 
    input [wrd_size-1:0] Alu_in1, Alu_in2; 
    input [sel_width-1:0] Alu_sel;         
    output reg [wrd_size-1:0] Alu_out;     
    output Alu_zero_flg;                   
    localparam NOP = 3'b000, 
               ADD = 3'b001, 
               SUB = 3'b010, 
               AND = 3'b011, 
               OR  = 3'b100, 
               SLT = 3'b101, 
               SRT = 3'b110, 
               NOT = 3'b111; 
    assign Alu_zero_flg = ~|Alu_out;
    always @(*) begin
        case(Alu_sel)
            NOP:  Alu_out = 0;           
            AND:  Alu_out = Alu_in1 & Alu_in2; 
            OR:   Alu_out = Alu_in1 | Alu_in2; 
            ADD:  Alu_out = Alu_in1 + Alu_in2; 
            SUB:  Alu_out = Alu_in1 - Alu_in2; 
            NOT:  Alu_out = ~Alu_in1;          
            SLT:  Alu_out = Alu_in1 << Alu_in2; 
            SRT:  Alu_out = Alu_in1 >> Alu_in2; 
            default: Alu_out = 0;          
        endcase
    end
endmodule
\end{lstlisting}

\newpage

\begin{lstlisting}[language=Verilog, basicstyle=\ttfamily\small, keywordstyle=\color{blue}\bfseries, stringstyle=\color{purple}, commentstyle=\color{gray}\itshape, numbers=none, showstringspaces=false, tabsize=4, breaklines=true, frame=single, caption={Example training sample for the functionality equivalence checking task.}, label=lst:fec_data_example, columns=fullflexible, keepspaces=true, xleftmargin=0pt, xrightmargin=0pt, aboveskip=0pt, belowskip=0pt]
Code:
module AND2_X4 (A1, A2, ZN);
    input A1;
    input A2;
    output ZN;
    and(ZN, A1, A2);
    specify
        (A1 => ZN) = (0.1, 0.1);
        (A2 => ZN) = (0.1, 0.1);
    endspecify
endmodule
\end{lstlisting}

\begin{lstlisting}[language=Verilog, basicstyle=\ttfamily\small, keywordstyle=\color{blue}\bfseries, stringstyle=\color{purple}, commentstyle=\color{gray}\itshape, numbers=none, showstringspaces=false, tabsize=4, breaklines=true, frame=single, columns=fullflexible, keepspaces=true, xleftmargin=0pt, xrightmargin=0pt, aboveskip=0pt]
Equal:
module AND2_X4 (A1, A2, ZN);
    input A1;
    input A2;
    output ZN;
    assign ZN = A1 & A2;
    specify
        (A1 => ZN) = (0.1, 0.1);
        (A2 => ZN) = (0.1, 0.1);
    endspecify
endmodule
\end{lstlisting}

\begin{lstlisting}[language=Verilog, basicstyle=\ttfamily\small, keywordstyle=\color{blue}\bfseries, stringstyle=\color{purple}, commentstyle=\color{gray}\itshape, numbers=none, showstringspaces=false, tabsize=4, breaklines=true, frame=single, columns=fullflexible, keepspaces=true, xleftmargin=0pt, xrightmargin=0pt, aboveskip=0pt]
Unequal:
module AND2_X4 (A1, A2, ZN);
    input A1;
    input A2;
    output ZN;
    wire nA1;
    wire nA2;
    wire nZN;
    nand(nZN, nA1, nA2);
    not(nA1, A1);
    not(nA2, A2);
    not(ZN, nZN);
    specify
        (A1 => ZN) = (0.1, 0.1);
        (A2 => ZN) = (0.1, 0.1);
    endspecify
endmodule
\end{lstlisting}

\end{document}

%% file: secs/0_abstract.tex
\begin{abstract}

Developing effective representations for register transfer level (RTL) designs is crucial for accelerating the hardware design workflow. Existing approaches, however, typically rely on a single data modality, either the RTL code or its associated graph-based representation, limiting the expressiveness and generalization ability of the learned representations. For RTL, the control data flow graph (CDFG) offers a comprehensive structural representation that preserves complete information, while the code modality explicitly encodes semantic and functional information. We argue that integrating these complementary modalities is essential for a thorough understanding of RTL designs. To this end, we propose UniRTL, a multimodal pretraining framework that learns unified RTL representations by jointly leveraging code and CDFG. UniRTL achieves fine-grained alignment between code and graph through mutual masked modeling and employs a hierarchical training strategy that incorporates a pretrained graph-aware tokenizer and staged alignment of text (\emph{i.e.}, functional summary) and code prior to graph integration. We evaluate UniRTL on two downstream tasks, performance prediction and code retrieval, under multiple settings. Experimental results show that UniRTL consistently outperforms prior methods, establishing it as a more robust and powerful foundation for advancing hardware design automation.

\end{abstract}

%% file: secs/1_introduction.tex
\section{Introduction}

Register transfer level (RTL) is a critical abstraction in the electronic design automation (EDA) workflow that describes the flow of data between registers and the logical operations performed on that data. 
As the front end of hardware design, deriving effective RTL representations can substantially accelerate the entire design process.
For instance, developing informative RTL representations for performance prediction enables hardware designers to obtain instant feedback on key quality metrics such as area and delay, bypassing the need for time-consuming logic synthesis~\citep{sengupta2022good, moravej2025graph, liu2025beyond}. 
Beyond performance prediction, effective RTL representations also facilitate tasks like code retrieval~\citep{liu2025deeprtl2}, which allows for the efficient identification and reuse of relevant design modules.
With the recent proliferation of large language models (LLMs) for RTL code generation~\citep{pei2024betterv, zhao2025codev, liu2025craftrtl, liudeeprtl}, the development of powerful representations for retrieval has become even more important.
These representations play a pivotal role in retrieval-augmented generation (RAG)~\citep{lewis2020retrieval}, thereby potentially enhancing the performance of RTL code generation systems.

Despite achieving promising performance, current approaches to RTL representation learning typically rely on a single data modality, either the RTL code or its associated graph-based representation, limiting the expressiveness and generalization ability of the learned representations. 
For example, in the context of performance prediction, VeriDistill~\citep{moravej2025graph} derives representations by feeding RTL code into LLMs specifically fine-tuned for RTL code generation and aggregating token-level embeddings for prediction.
On the other hand, StructRTL~\citep{liu2025beyond} constructs representations using a structure-aware self-supervised learning framework applied to the control data flow graph (CDFG) of RTL designs. 
Similarly, for the code retrieval tasks, DeepRTL2~\citep{liu2025deeprtl2} generates embeddings directly from RTL code using its backbone LLM.
While the code modality explicitly encodes semantic and functional information, the graph modality captures critical structural relationships that are often opaque from code. 
To achieve a more comprehensive understanding of RTL designs and obtain more robust and powerful representations, it is essential to develop methods that can effectively bridge these two modalities with complementary information.

In the software domain, GraphCodeBERT~\citep{guo2021graphcodebert} enhances code understanding by pretraining representations of programming languages with data flow information.
Despite its effectiveness, the model exhibits several notable limitations.
First, there is a weak alignment between code and data flow established by the variable-alignment task, which merely locates variable nodes in the code without capturing their full semantic relationships.
Second, the data flow representation itself is limited, as its nodes are restricted to variables, thereby overlooking other critical elements like operators and control flow, which are essential for tasks such as performance prediction and code retrieval.
Finally, the model directly feeds variable-level data flow nodes into a Transformer~\citep{vaswani2017attention} without employing a graph-aware tokenizer, which may hinder its ability to capture the nuanced and intricate structural relationships inherent in the graph.
Recently, CircuitFusion~\citep{fangcircuitfusion} has been proposed for constructing multimodal fused representations of RTL by incorporating code, structural graphs, and functional summaries. 
In contrast to GraphCodeBERT, which adopts a unified Transformer architecture, CircuitFusion first derives unimodal representations using three independent encoders, and subsequently integrates them through a cross-attention mechanism. 
Nevertheless, its alignment strategy remains coarse-grained, where it relies on contrastive learning between text-code and text-graph pairs while neglecting fine-grained alignment between code and graph—two modalities that contain more detailed and richer information.

To bridge this gap, we propose UniRTL, a novel multimodal pretraining framework that learns unified RTL representations by leveraging complementary modalities of RTL. 
UniRTL addresses the limitations of prior work by achieving fine-grained cross-modal alignment through mutual masked modeling. 
Following GraphCodeBERT~\cite{guo2021graphcodebert}, UniRTL employs a unified Transformer architecture to integrate different modalities, thereby eliminating the complexity of designing modality-specific encoders and enabling more seamless interaction across different modalities. 
Meanwhile, UniRTL adopts a hierarchical training strategy: a graph-aware tokenizer is first pretrained to enable the Transformer to better capture the nuanced structural dependencies in the graph, and alignment between text (\textit{i.e.}, functional summary) and code is performed before incorporating the graph, which maximizes data utilization given the greater availability of text-code pairs compared to graph data.
Moreover, instead of relying on data flow, UniRTL leverages CDFGs, which preserve complete information without loss and can be faithfully converted back to code.

We evaluate UniRTL on two downstream tasks, \textit{i.e.}, performance prediction and code retrieval, each under multiple settings. 
For performance prediction, we examine post-synthesis area and delay estimation both with and without the incorporation of netlist information, consistent with the setting of StructRTL~\cite{liu2025beyond}.
For code retrieval, we consider scenarios where the query is either text or code, following the setup of DeepRTL2~\cite{liu2025deeprtl2}.
Across all tasks and settings, UniRTL consistently outperforms prior methods, demonstrating the effectiveness 
of our framework.
The code for UniRTL is open-sourced at \url{https://github.com/cure-lab/UniRTL}.

%% file: secs/2_related_work.tex
\section{Related Works}

\textbf{RTL Representation Learning.}
Register transfer level (RTL) is a critical abstraction in the hardware design workflow, typically expressed in hardware description languages (HDLs) such as Verilog to specify data transfers between registers and the associated logical operations. 
Modern hardware design is inherently complex and involves multiple stages: natural language specifications are first manually translated into HDLs, which are then synthesized into circuit elements. 
Hardware designers often must wait for the time-consuming logic synthesis process to generate netlists and evaluate quality metrics, making iterative refinement slow and costly. To mitigate this bottleneck, prior research on RTL representation learning has primarily focused on performance prediction.
For example, \citet{sengupta2022good} employ a graph attention network (GAT)~\cite{velivckovic2018graph} on constructed CDFGs for delay and power prediction, while StructRTL~\cite{liu2025beyond} introduces a structure-aware self-supervised learning framework on CDFGs for post-synthesis area and delay prediction. 
VeriDistill~\cite{moravej2025graph}, in contrast, derives RTL representations using LLMs specifically fine-tuned for RTL code generation~\cite{pei2024betterv, cui2024origen, zhao2025codev, liu2025craftrtl, liudeeprtl}.
\begin{figure*}[t]
    \centering
    \includegraphics[width=0.79\linewidth]{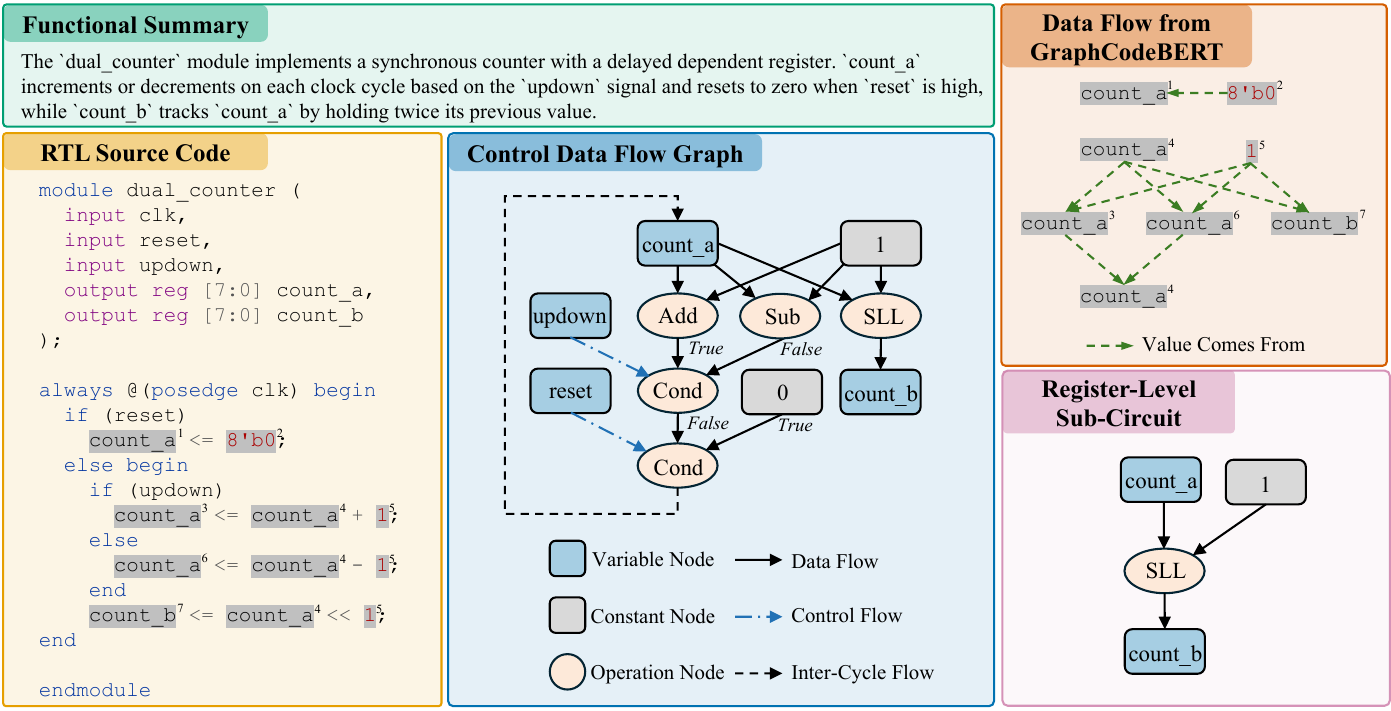}
    \caption{Example data point from our dataset, including RTL source code, and its corresponding functional summary and CDFG. For comparison, data flow~\citep{guo2021graphcodebert} and register-level sub-circuit~\citep{fangcircuitfusion} are also shown, demonstrating the completeness of the constructed CDFG.}
    \label{fig:sample_data}
\end{figure*}
Beyond performance prediction, DeepRTL2~\cite{liu2025deeprtl2} explores the task of code retrieval, motivated by the high reusability of hardware designs. It develops a versatile model capable of both generation- and embedding-based tasks, where text and code embeddings are obtained from the backbone LLM.
Despite these advances, existing approaches often rely on a single data modality, either the RTL code or its corresponding graph-based representation, which limits the expressiveness and generalization ability of the learned representations.

\textbf{Multimodal Representation Learning.}
Multimodal representation learning aims to learn joint representations from multiple modalities, with recent advances spanning a variety of domains, including vision-language~\citep{radford2021learning, bao2022vlmo, li2021align, li2022blip, li2023blip, jiangvlm2vec} and speech-text~\citep{chuang2020speechbert, tang2022unified, yu2023speech}.
By integrating complementary information across modalities, these approaches enable the development of more robust and powerful representations for a wide range of tasks.
Among existing works, the one most closely related to ours is GraphCodeBERT~\citep{guo2021graphcodebert}, which leverages data flow information to enhance code representation learning.
However, its alignment strategy is limited: it merely identifies variable nodes in the code without capturing their full semantic relationships.
Moreover, the employed data flow is incomplete, as it excludes critical elements such as operators and control flow, and the absence of a graph-aware tokenizer restricts the model's ability to capture the nuanced and intricate structural relationships inherent in the graph.
Another relevant effort is CircuitFusion~\citep{fangcircuitfusion}, which learns multimodal fused representations from RTL code, structural graphs, and functional summaries.
Nevertheless, its alignment strategy relies on coarse-grained contrastive learning between text–code and text–graph pairs, while overlooking fine-grained alignment between code and graph.
In addition, its dataset contains only 41 designs, and alignment is performed at the register sub-circuit level, which fails to capture the full semantics of entire modules or designs.
In contrast, UniRTL achieves fine-grained alignment between code and graph through mutual masked modeling and is pretrained on a large-scale dataset. Furthermore, the adopted CDFGs preserve complete information without loss and can be faithfully converted back to code.

%% file: secs/3_methodology.tex
\section{Methodology}

In this section, we detail the dataset construction process, with particular emphasis on CDFG conversion, and present the overall dataset statistics.
We then introduce the model architecture of UniRTL, highlighting both the mutual masked modeling alignment strategy and the hierarchical training strategy, in which a graph-aware tokenizer is first pretrained and text–code alignment is performed prior to incorporating the graph, thereby maximizing data utilization and enhancing model performance.

\begin{figure*}[t]
    \centering
    \includegraphics[width=0.79\linewidth]{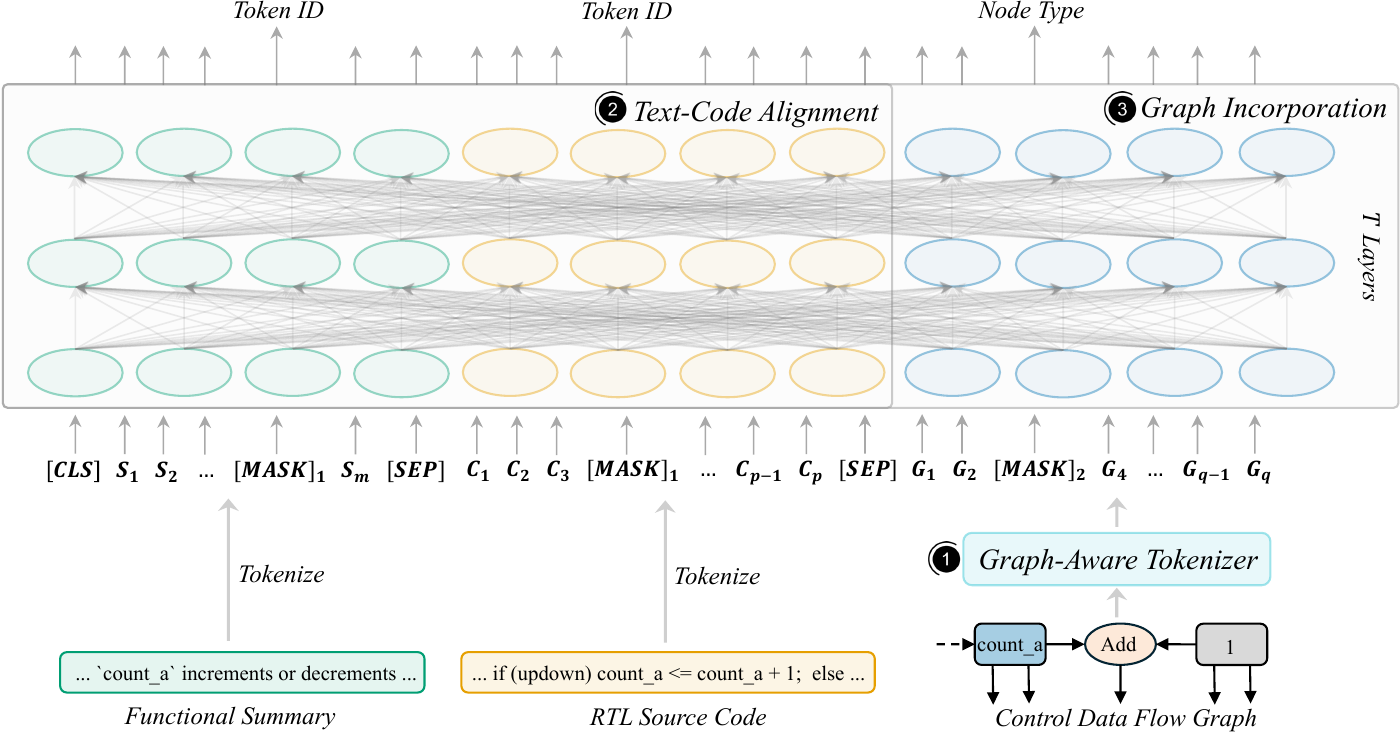}
    \caption{Overview of UniRTL. The framework achieves fine-grained cross-modal alignment via mutual masked modeling, and adopts a hierarchical training strategy: a graph-aware tokenizer is first pretrained, and text-code alignment is performed prior to graph incorporation.}
    \label{fig:overview}
\end{figure*}

\subsection{Dataset Construction}
\label{subsec:dataset_construction}
In this work, we collect datasets from multiple sources, including RTLCoder~\citep{liu2024rtlcoder}, MG-Verilog~\citep{zhang2024mg}, DeepRTL~\citep{liudeeprtl}, and DeepCircuitX~\citep{li2025deepcircuitx}. 
These datasets contain original RTL designs paired with their corresponding functional summaries. 
To construct CDFGs from RTL source code, we first compile the designs into RTL intermediate language (RTLIL) using Yosys~\citep{wolf2013yosys}, a simplified form that preserves semantic completeness while reducing designs to basic assignment and register-transfer operations, thereby simplifying CDFG extraction.
Next, we apply the Stagira Verilog parser~\cite{chen2023incremental} to generate an abstract syntax tree (AST) from the RTLIL, and then traverse the AST to extract the CDFG. An example data sample is shown in Figure~\ref{fig:sample_data}.
Note that not all collected RTL designs can be successfully converted into CDFGs, as many originate from open-source GitHub repositories or are generated by LLMs and may contain syntax errors leading to compilation failures.
Nevertheless, we retain these noisy samples for text–code alignment, enabling the model to learn more robust and generalizable representations while maximizing data utilization.
In total, our dataset contains 132,008 RTL designs, of which 38,888 are successfully converted into CDFGs.
Further analysis of the samples that fail to convert to CDFGs and the representativeness of the CDFG-convertible subset is provided in Appendix~\ref{appendix:cdfg_conversion_failure_analysis}.

\subsection{Model Architecture}
\label{sec:model_architecture}
We adopt a unified Transformer architecture as the backbone of UniRTL. Specifically, we use CodeBERT-base-mlm~\citep{feng2020codebert}\footnote{\url{https://huggingface.co/microsoft/codebert-base-mlm}} as our base model, pretrained on the CodeSearchNet~\citep{husain2019codesearchnet} code corpus using masked language modeling~\citep{devlin2019bert}. This pretrained model provides UniRTL with rich prior knowledge of code.
The overall framework of UniRTL is illustrated in Figure~\ref{fig:overview}. 
UniRTL achieves fine-grained cross-modal alignment through mutual masked modeling, especially for the code and graph.
Besides, to help the model better capture the nuanced and intricate structural relationships within the graph and maximize data utilization, we adopt a hierarchical training strategy, where a graph-aware tokenizer is first pretrained to encode structure-aware information in the CDFG, and text-code alignment is performed before the graph incorporation.

\textbf{Graph-Aware Tokenizer.}
Unlike GraphCodeBERT~\citep{guo2021graphcodebert}, which directly feeds flattened variable nodes from the data flow into the Transformer, we design a graph-aware tokenizer tailored to encode structure-aware information from the CDFG.
This enables the model to more effectively capture the nuanced and intricate structural relationships within the graph.
The graph-aware tokenizer combines a graph isomorphism network (GIN)~\cite{xupowerful} with a lightweight Transformer to jointly capture local structural dependencies and global contextual information.
Specifically, given a graph $\mathcal{G}=\{\mathbb{V}, \mathbb{E}\}$, where $\mathbb{V}$ denotes the set of nodes and $\mathbb{E}$ the set of edges, we encode each node $v_{i} \in \mathbb{V}$ as:
\begin{equation}
    \mathbf{H}_i = \text{one-hot}(\text{type}(v_i)) \parallel \text{width}(v_i) \parallel \text{pca}(\phi_{\text{text}}(\text{desc}(v_i)))
\end{equation}
This representation concatenates the one-hot encoding of the node type, the node width, and the embedding of its textual description. $\phi_{\text{text}}$ denotes the text encoder, for which we use all-mpnet-base-v2\footnote{\url{https://huggingface.co/sentence-transformers/all-mpnet-base-v2}}.
To balance the contribution of different components, we apply principal component analysis (PCA)~\citep{mackiewicz1993principal} to reduce the dimensionality of the description embedding from 768 to 32, matching the number of node types in our graphs.
Incorporating description embeddings proves particularly effective, as it facilitates information alignment between the graph and code.
After constructing the initial node embeddings, we feed the graph into a GIN to obtain node representations capturing local structural dependencies:
\begin{equation}
\mathbf{L}_{i}^{(k)} = \operatorname{MLP}^{(k)} \left( \left(1 + \epsilon^{(k)}\right) \cdot \mathbf{L}_{i}^{(k-1)} + \sum_{j \in \mathcal{N}(i)} \mathbf{L}_{j}^{(k-1)} \right)
\end{equation}
where $\mathbf{L}_{i}^{(0)}=\mathbf{H}_{i}$ is the initial embedding of node $v_{i}$, $\mathcal{N}(i)$ denotes the neighborhood of node $v_{i}$, and $\epsilon^{(k)}$ is a learnable scalar. After stacking $K$ GIN layers, we obtain the local structural embedding $\mathbf{L}_{i}=\mathbf{L}_{i}^{(K)}$.
To incorporate global contextual information across the entire graph, we further process the GIN embeddings with a lightweight Transformer encoder, which takes $\{\mathbf{L}_{i}\}_{i\in\mathbb{V}}$ as input and produces refined node embeddings $\{\mathbf{G}_{i}\}_{i\in\mathbb{V}}$. 
The graph-aware tokenizer is pretrained with two objectives, structure-aware masked node modeling and edge prediction, enabling it to encode nuanced and intricate structural relationships within the graph. The embeddings $\{\mathbf{G}_{i}\}_{i\in\mathbb{V}}$ then serve as the input to UniRTL.
For further details on the graph-aware tokenizer architecture and the pretraining tasks, please refer to Appendix~\ref{appendix:graph_aware_tokenizer_details}.


\textbf{Text-Code Alignment}.
Since text-code pairs are more abundant than graph data, we first perform text-code alignment prior to incorporating the graph.
This stage serves as a warm-up that familiarizes the model with RTL knowledge while maximizing data utilization.
The alignment is achieved through mutual masked modeling.
Specifically, the functional summary and RTL source code are tokenized into sequences $\{\mathbf{S}_{i}\}_{}$ and $\{\mathbf{C}_{i}\}$, respectively.
We then randomly mask 20\% of the tokens, with 80\% of the masked positions replaced by a special \texttt{[MASK]}$_{\texttt{1}}$ token, 10\% by a random token, and 10\% left unchanged.
UniRTL is pretrained to recover these masked tokens by predicting their original token IDs.
Since text and code encode complementary semantic information, masking one modality encourages the model to leverage the other for recovery, thereby promoting in-depth alignment between text and code.

\textbf{Graph Incorporation.}
After pretraining the graph-aware tokenizer and completing text-code alignment, we incorporate graph information into UniRTL to enable fine-grained alignment between code and graph.
Specifically, given a graph, we first process it with the graph-aware tokenizer to obtain node embeddings $\{\mathbf{G}_{i}\}_{i\in\mathbb{V}}$ that capture the nuanced and intricate structural relationships within the graph.
These embeddings are then fed into UniRTL, where alignment with text and code is achieved through mutual masked modeling.
For text and code, we follow the same masking strategy used in text-code alignment. For the graph, 20\% of the nodes are randomly selected and replaced with a learnable \texttt{[MASK]}$_{\texttt{2}}$ embedding. 
UniRTL is trained to recover the masked nodes by predicting their original node types, while simultaneously recovering masked text and code tokens.
This joint objective encourages UniRTL to capture the full semantic relationships between code and graph.
To preserve the graph's topological structure, we augment $\{\mathbf{G}_{i}\}_{i\in\mathbb{V}}$ with global positional encodings $\{\mathbf{P}_{i}\}_{i\in\mathbb{V}}$~\citep{rampavsek2022recipe} before feeding them into UniRTL. 
The global positional encodings are derived from the eigenvectors of the symmetric normalized graph Laplacian~\citep{chung1997spectral}:
\begin{equation}
    L = I - D_{\text{in}}^{-1/2}(\frac{A+A^\mathrm{T}}{2})D_{\text{out}}^{-1/2}
\end{equation}
where $A$ is the adjacency matrix, and $D_{\text{in}}$ and $D_{\text{out}}$ denote the in-degree and out-degree matrices, respectively. The eigenvalues and eigenvectors of $L$ are computed by solving:
\begin{equation}
    L \mathbf{x} = \lambda \mathbf{x}
\end{equation}
where $\{\lambda_{i}\}$ are the eigenvalues and $\{\mathbf{x}_{i}\}$ are the corresponding eigenvectors. We select the 16 smallest eigenvalues and their associated eigenvectors to construct the global positional encodings.
Before integrating $\{\mathbf{P}_{i}\}_{i\in\mathbb{V}}$ with $\{\mathbf{G}_{i}\}_{i\in\mathbb{V}}$, a linear projection layer is applied to map the positional encodings to the same dimensionality as the node embeddings.
Finally, an adapter is employed to project $\{\mathbf{G}_{i}\}_{i\in\mathbb{V}}$ into the joint text-code embedding space, thereby facilitating more effective cross-modal alignment. 

%% file: secs/4_experiment.tex
\section{Experimental Results}

In this section, we elaborate the experimental settings and present the results. We evaluate UniRTL on two downstream tasks, performance prediction and code retrieval, each under multiple settings.
For performance prediction, we examine post-synthesis area and delay estimation, both with and without the incorporation of netlist information.
For code retrieval, we consider scenarios where the query is either text or code.
Across all tasks and settings, UniRTL consistently outperforms baseline methods, demonstrating the robustness and effectiveness of our framework.

\input{tabs/performance_prediction_without_netlist}

\subsection{Baseline Methods}
For performance prediction, we consider several baselines: StructRTL~\citep{liu2025beyond}, VeriDistill~\citep{moravej2025graph}, and DeepRTL2~\citep{liu2025deeprtl2}.
StructRTL derives RTL representations through a structure-aware self-supervised learning framework on CDFGs, while VeriDistill and DeepRTL2 obtain RTL representations by leveraging LLMs fine-tuned for RTL code generation to produce token-level embeddings, which are subsequently aggregated via mean or max pooling for prediction.
Particularly, VeriDistill adopts the open-source Verilog LLM CodeV~\citep{zhao2025codev}, which offers three variants: CodeV-DS-6.7B, CodeV-CL-7B, and CodeV-QW-7B, fine-tuned from DeepSeek-Coder~\citep{guo2024deepseek}, CodeLlama~\cite{roziere2023code}, and CodeQwen~\citep{bai2023qwen}, respectively.
DeepRTL2 provides two variants, fine-tuned from Llama-3.1~\citep{grattafiori2024llama} and DeepSeek-Coder, respectively.
We include all these variants in our comparison.
In addition, we evaluate an end-to-end prediction method that employs a GAT directly over CDFGs for performance estimation~\citep{sengupta2022good}.
For code retrieval, we compare against state-of-the-art general-purpose text embedding models, including OpenAI's text-embedding-3-small and text-embedding-3-large~\citep{neelakantan2022text}, NV-Embed-v2~\citep{lee2025nv} and GritLM-7B~\citep{muennighoff2025generative}, as well as customized RTL embedding models (DeepRTL2-Llama and DeepRTL2-DeepSeek).
We also incorporate GraphCodeBERT~\citep{guo2021graphcodebert} and CircuitFusion~\citep{fangcircuitfusion} as baselines for both tasks to highlight the necessity of our designs, including the use of complete graphs, the graph-aware tokenizer, and fine-grained alignment between code and graph. Importantly, we do not use GraphCodeBERT in its original, software-oriented form, but instead fine-tune it on the same RTL datasets and downstream tasks as UniRTL to ensure a fair comparison. For CircuitFusion, we reproduce its pretraining process using our CDFGs for fair comparison.
A detailed efficiency comparison, including pretraining cost, downstream fine-tuning cost, and inference memory usage, is provided in Appendix~\ref{appendix:efficiency_comparison}.

\subsection{Experimental Setup}
\label{subsec:experimental_settings}
In this subsection, we detail the hyperparameter configurations for the model architecture and training process. 
UniRTL adopts the same architecture as its base model, CodeBERT-base-mlm~\citep{feng2020codebert}, consisting of 12 Transformer layers with 12 attention heads per layer.
During the text–code alignment stage, the base model is trained for 5 epochs on 4 NVIDIA L40 GPUs with a per-device batch size of 32.
Training is performed using the AdamW optimizer~\citep{loshchilovdecoupled} with a learning rate of 8e-5 and a weight decay of 0.01.
To improve training stability, we employ a cosine learning rate scheduler with a warmup ratio of 0.03 and set the gradient accumulation steps to 8.
After graph incorporation, the model is further trained for 300 epochs on 2 NVIDIA L40 GPUs with a per-device batch size of 16. All other hyperparameter settings remain the same as in the text–code alignment stage.

\input{tabs/performance_prediction_with_netlist}





\subsection{Performance Prediction}
The experimental settings for performance prediction mainly follows StructRTL~\citep{liu2025beyond}.
Specifically, we predict post-synthesis area and delay values, where RTL designs are synthesized and mapped to post-mapping netlists using Yosys~\citep{wolf2013yosys} and ABC~\citep{brayton2010abc} with the SkyWater 130nm technology library~\citep{edwards2020google}. The area and delay values are then extracted from the generated netlists.
For fine-tuning, we adopt the dataset from StructRTL, which consists of 13,200 designs split into training and validation sets with an 0.8:0.2 ratio.
The task is formulated as a regression problem. 
After obtaining RTL representations with different methods, we fine-tune a three-layer multi-layer perceptron (MLP) to perform performance estimation.
Additional details of the fine-tuning process are provided in Appendix~\ref{appendix:performance_prediction_details}.
For evaluation, we report four standard regression metrics: mean absolute error (MAE), mean absolute percentage error (MAPE), coefficient of determination ($R^{2}$), and root relative squared error (RRSE). 
Detailed definitions of these metrics are provided in Appendix~\ref{appendix:evaluation_metrics_for_performance_prediction}.

The performance comparison of different methods are presented in Table~\ref{tab:performance_prediction_without_netlist}. 
Notably, UniRTL consistently outperforms all baselines across all evaluation metrics for both post-synthesis area and delay prediction, establishing a new state of the art.
Among the baselines, StructRTL achieves the strongest performance, highlighting the advantage of leveraging CDFGs over RTL source code, as CDFGs capture richer structural information that is critical for accurate performance estimation.
In contrast, GraphCodeBERT, despite incorporating data flow information, performs significantly worse than other methods.
This underperformance can be attributed to the limited scope of the data flow information it encodes, which is insufficient for this task, as well as its relatively small model size compared to LLM-based methods, resulting in weaker code embeddings.
Importantly, UniRTL, with a model size comparable to GraphCodeBERT, surpasses not only GraphCodeBERT but also much larger LLM-based methods, underscoring the effectiveness and efficiency of our framework.
Additionally, we conduct an ablation study by removing the code and graph components of UniRTL, yielding two variants: UniRTL (w/o code) and UniRTL (w/o graph), respectively.
We find that removing the graph component substantially degrades performance, underscoring the essential role of structural information encoded in CDFGs for performance prediction, while removing the code component results in a slight performance drop, indicating that code still provides complementary information that can enhance performance prediction.

To further enhance performance prediction, VeriDistill~\citep{moravej2025graph} and StructRTL~\citep{liu2025beyond} adopt a knowledge distillation strategy that transfers low-level insights from netlists into the performance predictor, \emph{i.e.}, the three-layer MLP. 
Following StructRTL, we collect synthesized post-mapping (PM) netlists and train a GIN to directly predict performance metrics from these netlists. 
Since the area and delay values are directly extracted from the PM netlists, this PM predictor achieves high accuracy and serves as the teacher model.
We then freeze the PM predictor and incorporate a knowledge distillation loss during the fine-tuning of the three-layer MLP, enabling it to integrate low-level information from the netlists.
Experimental results with the incorporation of netlist information are reported in Table~\ref{tab:performance_prediction_with_netlist}.
As shown, incorporating netlist information improves the performance of all methods. Nevertheless, UniRTL achieves state-of-the-art performance by surpassing all baselines across all evaluation metrics for both area and delay prediction, further demonstrating the robustness of our framework.
For additional details on the knowledge distillation process, please refer to Appendix~\ref{appendix:performance_prediction_with_netlist_info}.
We further discuss few-shot fine-tuning for performance prediction in Appendix~\ref{appendix:few_shot_finetuning_performance_prediction}.

\input{tabs/natural_language_code_search}

\input{tabs/functionality_equivalence_checking}

\subsection{Code Retrieval}
For code retrieval, we consider two scenarios in which the query is either text or code. 
Specifically, we adopt the settings of DeepRTL2~\citep{liu2025deeprtl2}, corresponding to its natural language code search and functionality equivalence checking tasks, respectively.

\textbf{Natural Language Code Search.}
Natural language code search aims to retrieve relevant code snippets from a large codebase given natural language queries. We formulate it as a retrieval problem using the bitext mining setting of the MTEB benchmark~\citep{muennighoff2023mteb}.
Specifically, the input for this task consists of a tuple $(\mathcal{S}, \mathcal{R})$, where $\mathcal{S}$ denotes a list of functional summaries in natural language and $\mathcal{R}$ the corresponding RTL designs. In this work, elements of $\mathcal{R}$ may be provided either as RTL code alone or as ``code \& graph'', where each RTL design includes both the code and its associated CDFG.
During evaluation, all queries $\{\mathcal{S}_i\}$ and candidates $\{\mathcal{R}_i\}$ are embedded into fixed-length vectors. For each query $\mathcal{S}_i$, cosine similarity is computed against all candidates, and the index $j=\underset{k}{\arg\max}\operatorname{cos}(\mathcal{S}_i, \mathcal{R}_k)$ is selected.
The retrieved $\mathcal{R}_j$ is regarded as the prediction for $\mathcal{S}_i$, while the corresponding $\mathcal{R}_i$ serves as the ground truth.
For training and evaluation, we use the dataset and benchmark provided by DeepRTL2, with the modification that designs failing to convert successfully into CDFGs are removed to ensure fairness. 
We adopt three evaluation metrics: Precision, Recall, and F1, with F1 serving as the main metric.
Further details of the experimental setup for this task are provided in Appendix~\ref{appendix:nlcs_experimental_setup}.

The experimental results are presented in Table ~\ref{tab:natural_language_code_search_result}. UniRTL consistently outperforms all baseline methods across all evaluation metrics, demonstrating the effectiveness of our framework. 
When restricted to the code-only format (UniRTL w/o graph), performance significantly degrades, highlighting the importance of incorporating graph information.
Furthermore, UniRTL's improvements over GraphCodeBERT and CircuitFusion demonstrate the benefits of our fine-grained cross-modal alignment, hierarchical training strategy, and the integration of complete graph information.
Interestingly, GraphCodeBERT even underperforms the variant of UniRTL where no graph is incorporated, which we hypothesize may be due to its targeted variable-alignment task interfering with the alignment between text and code, thereby hindering performance on natural language code search.

\textbf{Functionality Equivalence Checking.}
Functionality equivalence checking aims to determine whether two different RTL implementations exhibit identical behavior despite structural differences. This task follows the pair classification setting of the MTEB benchmark. 
Specifically, the input for this task consists of $N$ pairs of RTL designs, $\{(\mathcal{R}_{1}^{(1)},\mathcal{R}_{1}^{(2)})\}_{i=1}^{N}$, where each design can be represented either as code alone or as ``code \& graph''.
For each pair $(\mathcal{R}_{1}^{(1)},\mathcal{R}_{1}^{(2)})$, the model is expected to determine whether they are functionally equivalent by calculating the cosine similarity between their embedding vectors.
For training and evaluation, we adopt the dataset and benchmark provided by DeepRTL2, excluding designs that cannot be successfully converted to CDFGs to ensure fair evaluation.
We report five evaluation metrics for this task: Average Precision (AP), Accuracy, F1, Precision, and Recall, with AP serving as the main metric.
Further details on the experimental setup for this task are provided in Appendix~\ref{appendix:fec_experimental_setup}.

\begin{figure*}[t]
    \centering
    \includegraphics[width=0.9\linewidth]{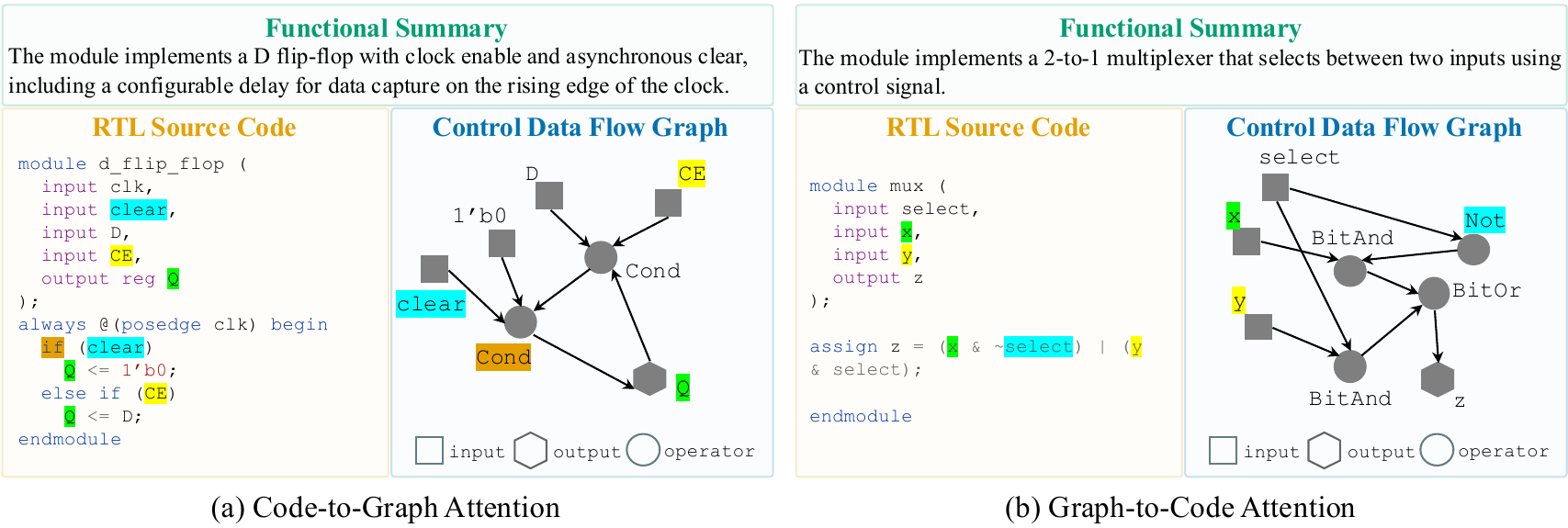}
    \caption{Qualitative analysis of code–graph relationships learned by mutual masked modeling. (a) Nodes with the highest attention scores from code tokens. (b) Code tokens with the highest attention scores from graph nodes. Corresponding code token–graph node pairs are highlighted with the same background color.}
    \label{fig:qualitative_analysis}
\end{figure*}

The performance comparison of different methods on the functionality equivalence checking task is presented in Table~\ref{tab:functionality_equivalence_checking_result}. UniRTL significantly outperforms all baseline methods on the main evaluation metric, further demonstrating the effectiveness of our framework. 
Removing the graph component (UniRTL w/o graph) leads to a substantial performance degradation, highlighting the importance of graph incorporation. 
Moreover, GraphCodeBERT performs better than the variant of UniRTL where no graph is incorporated, indicating that incorporating the data flow information can enhance the performance of functionality equivalence checking.
However, UniRTL's superior performance over GraphCodeBERT and CircuitFusion demonstrates that merely leveraging graph information or coarse multimodal fusion is insufficient; instead, dedicated strategies are essential to integrate the complete graph information, further validating the contributions of the various components in our framework. Additionally, we provide an ablation study of code–graph alignment strategies in Appendix~\ref{appendix:ablation_study_of_code_graph_alignment_strategies}.

\subsection{Qualitative Analysis of Code–Graph Relationships}
\label{subsec:qualitative_analysis_of_code_graph_relationships}
During the graph incorporation stage, the mutual masked prediction task requires the model to recover masked code tokens from the functional summary, surrounding code tokens, and graph nodes, and symmetrically to infer masked graph node types from the functional summary, neighboring graph nodes, and code tokens. To illustrate the nature of the resulting cross-modal alignment, we conduct a qualitative analysis of the code–graph relationships learned by the model.
Specifically, we analyze cross-modal attention patterns in the pretrained UniRTL model. For each input triplet (functional summary, RTL source code, CDFG), we extract attention scores by summing the attention weights across all heads in the last Transformer layer. We then examine attention in a bilateral manner: (1) code-to-graph, where for each code token we identify the graph nodes receiving the highest attention from that token; and (2) graph-to-code, where for each graph node we identify the code tokens receiving the highest attention from that node.

Figure~\ref{fig:qualitative_analysis} shows representative examples. In Figure~\ref{fig:qualitative_analysis}(a), the code tokens corresponding to \texttt{clear}, \texttt{CE}, and \texttt{Q} attend strongly to their respective nodes in the CDFG. In Figure~\ref{fig:qualitative_analysis}(b), the graph nodes \texttt{x} and \texttt{y} direct most of their attention to the corresponding code tokens \texttt{x} and \texttt{y}. These visualizations indicate that UniRTL learns to focus on semantically corresponding elements across modalities, thereby capturing meaningful alignments between specific code expressions and their associated CDFG nodes.


%% file: tabs/performance_prediction_without_netlist.tex
\begin{table*}[t]
\centering
\caption{Performance comparison of different methods on performance prediction tasks without the incorporation of netlist information. The best results are highlighted in bold.}
\label{tab:performance_prediction_without_netlist}
\setlength{\tabcolsep}{4pt}          
\begin{tabular}{lcccccccc}
\toprule
\multirow{2}{*}{w/o Netlist Info} & \multicolumn{4}{c}{\textbf{Area}} & \multicolumn{4}{c}{\textbf{Delay}} \\
\cmidrule(lr){2-5}\cmidrule(lr){6-9} 
 & MAE$\downarrow$ & MAPE$\downarrow$ & $R^{2}$$\uparrow$ & RRSE$\downarrow$ & MAE$\downarrow$ & MAPE$\downarrow$ & $R^{2}$$\uparrow$ & RRSE$\downarrow$ \\
\midrule
GAT     & 0.5497 & 0.09 & 0.5857 & 0.6437 & 0.7327 & 0.13 & 0.6639 & 0.5797 \\
StructRTL     & 0.3649 & \textbf{0.06} & 0.7463 & 0.5037 & 0.5414 & 0.10 & 0.7630 & 0.4868 \\
CodeV-DS-6.7B & 0.8967 & 0.17 & 0.4862 & 0.6973 & 0.6403 & 0.12 & 0.3905 & 0.7807 \\
CodeV-CL-7B   & 0.7982 & 0.15 & 0.5755 & 0.6515 & 0.5620 & 0.10 & 0.5174 & 0.6947 \\
CodeV-QW-7B   & 0.7229 & 0.13 & 0.6353 & 0.6039 & 0.5340 & 0.09 & 0.5277 & 0.6872 \\
DeepRTL2-Llama   & 0.6988 & 0.12 & 0.6758 & 0.5694 & 0.5756 & 0.10 & 0.5017 & 0.7059 \\
DeepRTL2-DeepSeek  & 0.7802 & 0.14 & 0.6225 & 0.6144 & 0.6357 & 0.11 & 0.4137 & 0.7657 \\
GraphCodeBERT   & 0.8424 & 0.15 & 0.5207 & 0.6923 & 0.6109 & 0.11 & 0.3989 & 0.7753 \\
CircuitFusion   & 0.7762 & 0.14 & 0.6175 & 0.6185 & 0.5272 & 0.09 & 0.5619 & 0.6619 \\

\midrule
UniRTL     & \textbf{0.3510} & \textbf{0.06} & \textbf{0.7682} & \textbf{0.4815} & \textbf{0.3384} & \textbf{0.06} & \textbf{0.7832} & \textbf{0.4656} \\
UniRTL (w/o code)     & 0.3671 & 0.07 & 0.7546 & 0.4954 & 0.3584 & 0.06 & 0.7602 & 0.4897 \\
UniRTL (w/o graph)    & 0.8818 & 0.15 & 0.5173 & 0.6948 & 0.6375 & 0.11 & 0.3839 & 0.7849 \\
\bottomrule
\end{tabular}
\end{table*}

%% file: tabs/performance_prediction_with_netlist.tex
\begin{table*}[t]
\centering
\caption{Performance comparison of different methods with the incorporation of netlist information. For reference, we also report the performance of the teacher model. The best results, excluding the teacher model, are highlighted in bold.}
\label{tab:performance_prediction_with_netlist}
\setlength{\tabcolsep}{4pt}          
\begin{tabular}{lcccccccc}
\toprule
\multirow{2}{*}{w/ Netlist Info} & \multicolumn{4}{c}{\textbf{Area}} & \multicolumn{4}{c}{\textbf{Delay}} \\
\cmidrule(lr){2-5}\cmidrule(lr){6-9} 
 & MAE$\downarrow$ & MAPE$\downarrow$ & $R^{2}$$\uparrow$ & RRSE$\downarrow$ & MAE$\downarrow$ & MAPE$\downarrow$ & $R^{2}$$\uparrow$ & RRSE$\downarrow$ \\
\midrule
PM Predictor & 0.2982 & 0.05 & 0.9334 & 0.2581 & 0.1688 & 0.03 & 0.9484 & 0.2272 \\
\midrule
GAT     & 0.4689 & 0.09 & 0.7954 & 0.4523 & 0.2926 & 0.05 & 0.8113 & 0.4344 \\
StructRTL                           & 0.3856 & 0.07 & 0.8676 & 0.3639 & 0.2381 & \textbf{0.04} & 0.8872 & 0.3359 \\
CodeV-DS-6.7B & 0.4896 & 0.09 & 0.7928 & 0.4552 & 0.3787 & 0.07 & 0.7235 & 0.5258 \\
CodeV-CL-7B   & 0.4192 & 0.08 & 0.8225 & 0.4213 & 0.3208 & 0.06 & 0.7696 & 0.4800 \\
CodeV-QW-7B   & 0.4397 & 0.08 & 0.8174 & 0.4273 & 0.3284 & 0.06 & 0.7687 & 0.4809 \\
DeepRTL2-Llama   & 0.4540 & 0.08 & 0.8332 & 0.4085 & 0.3707 & 0.07 & 0.7445 & 0.5054 \\
DeepRTL2-DeepSeek   & 0.4915 & 0.09 & 0.8287 & 0.4139 & 0.4014 & 0.07 & 0.7273 & 0.5222 \\
GraphCodeBERT   & 0.6008 & 0.11 & 0.7578 & 0.4922 & 0.4289 & 0.07 & 0.6907 & 0.5561 \\
CircuitFusion   & 0.5854 & 0.11 & 0.7902 & 0.4581 & 0.4341 & 0.08 & 0.7143 & 0.5345 \\
\midrule
UniRTL   & \textbf{0.3362} & \textbf{0.06} & \textbf{0.8879} & \textbf{0.3349} & \textbf{0.2302} & \textbf{0.04} & \textbf{0.8946} & \textbf{0.3247} \\
UniRTL (w/o code)   & 0.3462 & \textbf{0.06} & 0.8741 & 0.3548 & 0.2764 & 0.05 & 0.8817 & 0.3439 \\
UniRTL (w/o graph)   & 0.6121 & 0.11 & 0.7547 & 0.4953 & 0.4478 & 0.08 & 0.6775 & 0.5679 \\
\bottomrule
\end{tabular}
\end{table*}

%% file: tabs/natural_language_code_search.tex
\begin{table*}[t]
\centering
\caption{Performance comparison of different methods on the natural language code search task, with F1 used as the main metric. The best scores are highlighted in bold.}
\label{tab:natural_language_code_search_result}
\begin{tabular}{lcccc}
\toprule
Model            & Design Format & Precision$\uparrow$        & Recall$\uparrow$        & F1$\uparrow$ (Main)                    \\ \hline
text-embedding-3-small          &   code   &   0.254   &    0.350       &   0.277                            \\
text-embedding-3-large           &  code   &   0.350    &   0.442        &  0.375                             \\ 
GritLM-7B       &  code   &   0.393    &   0.475   &   0.414           \\ 
NV-Embed-v2     &  code   &   0.367    &   0.450   &   0.389             \\ 
DeepRTL2-Llama  &  code   &   0.557    &   0.608   &   0.572           \\ 
DeepRTL2-DeepSeek & code  &   0.532    &   0.592   &   0.547   \\ 
GraphCodeBERT     & code \& graph &   0.616  &  0.675   &   0.634    \\
CircuitFusion     & code \& graph &   0.542  &  0.608   &   0.560    \\ \hline
UniRTL  & code \& graph  &    \textbf{0.650}    &   \textbf{0.692}   &   \textbf{0.662}   \\
UniRTL (w/o graph) & code &   0.630    &   0.683   &   0.644  \\ 
\bottomrule
\end{tabular}
\end{table*}

%% file: tabs/functionality_equivalence_checking.tex
\begin{table*}[t]
\centering
\caption{Performance comparison of different methods on the functionality equivalence checking task, with average precision (AP) as the main metric. The best results are highlighted in bold.}
\label{tab:functionality_equivalence_checking_result}
\setlength{\tabcolsep}{4pt}
\begin{tabular}{lcccccc}
\toprule
Model            & Design Format & AP$\uparrow$ (Main) &  Accuracy$\uparrow$ & F1$\uparrow$ & Precision$\uparrow$ & Recall$\uparrow$               \\ \hline
text-embedding-3-small          &   code   &  0.543    &   0.613  &  0.696 &  0.545  &  0.960                \\
text-embedding-3-large           &  code   &  0.564    &   0.587  &  0.687 &      0.553  &  0.907                 \\ 
GritLM-7B       &  code   &   0.599    &  0.640    &  0.724    &  0.587   & 0.947  \\ 
NV-Embed-v2     &  code   &   0.554    &  0.607    &  0.667    &  0.547   & 0.853   \\ 
DeepRTL2-Llama  &  code   &  0.646     &  0.695    &  0.737    & 0.597    & 0.964 \\ 
DeepRTL2-DeepSeek & code  &  0.631     &  0.640    &  0.729    & 0.587    & 0.960 \\ 
GraphCodeBERT     & code \& graph &  0.730   &  0.733   &  \textbf{0.753}   &  0.613   & \textbf{0.973}  \\
CircuitFusion     & code \& graph &  0.628   &  0.667   &  0.736   &  0.602   & 0.947  \\ \hline
UniRTL  & code \& graph  &  \textbf{0.745}  &  \textbf{0.747}    &  \textbf{0.753}  &  \textbf{0.734}   & 0.773 \\
UniRTL (w/o graph) & code &  0.712 &  0.667    &  0.717  &  0.577   & 0.947 \\ 
\bottomrule
\end{tabular}
\vspace{-6pt}
\end{table*}

%% file: secs/5_limitations.tex
\section{Limitations}

Although UniRTL demonstrates strong performance, several limitations remain. It relies on successful CDFG conversion to fully exploit the graph modality, so designs that fail parsing or compilation cannot be used for graph incorporation, even though we retain them for text--code alignment and analyze the representativeness of the convertible subset. Our current implementation is also limited to Verilog, the amount of graph-paired pretraining data is smaller than the full text--code corpus, and the evaluation focuses on performance prediction and code retrieval. Extending UniRTL to more hardware description languages, larger graph-paired corpora, broader RTL understanding tasks, and substantially larger industrial designs remains important future work, especially because long RTL sequences and large CDFGs may introduce scalability challenges. In addition, our experiments are conducted under fixed synthesis settings, and further studies across more diverse synthesis flows would help assess robustness in practical deployment scenarios.

%% file: secs/6_conclusion.tex
\section{Conclusion}
In this work, we introduce UniRTL, a multimodal pretraining framework that unifies RTL code and CDFGs for robust RTL representation learning.
Unlike prior approaches that rely on simplified data flows or register-level sub-circuits, UniRTL leverages CDFGs that preserve complete design information and can be faithfully converted back to code.
Furthermore, instead of establishing only weak code-graph alignment through contrastive objectives, UniRTL achieves fine-grained cross-modal alignment through mutual masked modeling.
To better capture the nuanced and intricate structural dependencies within graphs, UniRTL employs a hierarchical training strategy: a graph-aware tokenizer is first pretrained, and text–code alignment is performed as a warm-up stage to maximize data utilization before incorporating the graph.
We evaluate UniRTL on two downstream tasks, performance prediction and code retrieval, each under multiple settings. 
Experimental results demonstrate that UniRTL consistently outperforms previous methods across all tasks and settings, validating its robustness and effectiveness.
Overall, UniRTL establishes a more general and powerful foundation for advancing hardware design automation.

%% file: tabs/training_time_comparison.tex
\begin{table*}[t]
\centering
\caption{Efficiency comparison of the baselines and UniRTL. We separately report model initialization, pretraining cost, downstream fine-tuning cost, and inference memory usage.}
\label{tab:training_time_comparison}
\scriptsize
\setlength{\tabcolsep}{3pt}
\resizebox{\textwidth}{!}{%
\begin{tabular}{llcclcl}
\toprule
Model & Initialization & Pretraining time & Pretraining hardware & Fine-tuning time & Fine-tuning hardware & Inference memory (bfloat16) \\
\midrule
VeriDistill & CodeV & None & None & 12 h & 8$\times$ NVIDIA V100 & $\sim$14 GB \\
DeepRTL2 & Llama/DeepSeek & 70 h & 8$\times$ NVIDIA A800 & None & None & $\sim$15 GB \\
GraphCodeBERT & CodeBERT & 83 h & 16$\times$ NVIDIA V100 & $<$1 h & 1$\times$ NVIDIA L40 & $\sim$250 MB \\
GritLM-7B & Mistral-7B & 48 h & 64$\times$ NVIDIA A100 & None & None & $\sim$14 GB \\
text-embedding-3-small / -large & N/A & N/A & N/A & None & None & API-only inference \\
NV-Embed-v2 & Mistral-7B & N/A & N/A & None & None & $\sim$15 GB \\
GAT (graph-only baseline) & None & None & None & 1 h & 1$\times$ NVIDIA L40 & $\sim$12 MB \\
StructRTL & None & 40 h & 1$\times$ NVIDIA L40 & $<$1 h & 1$\times$ NVIDIA L40 & $\sim$50 MB \\
UniRTL & CodeBERT & 45 h & 2$\times$ NVIDIA L40 & $<$1 h & 1$\times$ NVIDIA L40 & $\sim$300 MB \\
\bottomrule
\end{tabular}
}
\end{table*}

%% file: tabs/nlcs_hyperparameters.tex
\begin{table*}[t]
\centering   
\caption{Hyperparameter configurations employed during fine-tuning for code retrieval tasks.}
\label{tab:two_task_hyperparams}
\begin{subtable}{0.49\textwidth}
  \centering
  \begin{tabular}{@{} l c @{}}
    \toprule
    Hyperparameter & Value \\
    \midrule
    finetuning\_type & full \\
    temperature & 0.05 \\
    normalize & true \\
    optimizer & AdamW \\
    learning\_rate & 5e-5 \\
    weight\_decay & 0.01 \\
    batch\_size & 64 \\
    epochs & 8 \\
    lr\_scheduler\_type & cosine \\
    warmup\_ratio & 0.03 \\
    gradient\_accumulation\_steps & 8 \\
    \bottomrule
  \end{tabular}
  \caption{Natural language code search}
  \label{tab:nlcs_hyperparameters}
\end{subtable}%
\hfill
\begin{subtable}{0.49\textwidth}
  \centering
  \begin{tabular}{@{} l c @{}}
    \toprule
    Hyperparameter & Value \\
    \midrule
    finetuning\_type & full \\
    temperature & 0.05 \\
    normalize & true \\
    optimizer & AdamW \\
    learning\_rate & 5e-5 \\
    weight\_decay & 0.01 \\
    batch\_size & 16 \\
    epochs & 16 \\
    lr\_scheduler\_type & cosine \\
    warmup\_ratio & 0.03 \\
    gradient\_accumulation\_steps & 8 \\
    max\_hard\_negatives & 3 \\
    \bottomrule
  \end{tabular}
  \caption{Functionality equivalence checking}
  \label{tab:fec_hyperparameters}
\end{subtable}
\vspace{-20pt}
\end{table*}





%% file: tabs/few_shot_performance_prediction.tex
\begin{table}[t]
\centering
\caption{Few-shot performance prediction results of UniRTL without netlist information.}
\label{tab:few_shot_performance_prediction}
\setlength{\tabcolsep}{4pt}
\begin{tabular}{lcccccccc}
\toprule
\multirow{2}{*}{Training data} & \multicolumn{4}{c}{\textbf{Area}} & \multicolumn{4}{c}{\textbf{Delay}} \\
\cmidrule(lr){2-5}\cmidrule(lr){6-9}
& MAE$\downarrow$ & MAPE$\downarrow$ & $R^{2}$$\uparrow$ & RRSE$\downarrow$ & MAE$\downarrow$ & MAPE$\downarrow$ & $R^{2}$$\uparrow$ & RRSE$\downarrow$ \\
\midrule
5\% & 0.8681 & 0.14 & -0.0604 & 1.0298 & 1.1069 & 0.18 & 0.2350 & 0.8746 \\
10\% & 0.5800 & 0.10 & 0.4334 & 0.7527 & 0.8273 & 0.15 & 0.5134 & 0.6976 \\
20\% & 0.4973 & 0.09 & 0.5630 & 0.6611 & 0.7352 & 0.13 & 0.6084 & 0.6258 \\
100\% & 0.3510 & 0.06 & 0.7682 & 0.4815 & 0.3384 & 0.06 & 0.7832 & 0.4656 \\
\bottomrule
\end{tabular}
\end{table}

%% file: tabs/incomplete_pretrain_performance_prediction.tex
\begin{table}[t]
\centering
\caption{Performance comparison of UniRTL (w/o graph) and CodeBERT-based baselines on performance prediction tasks without incorporating netlist information. CodeBERT-IP denotes the CodeBERT (Incomplete Pretrain).}
\label{tab:incomplete_pretrain_performance_prediction}
\setlength{\tabcolsep}{4pt}
\begin{tabular}{lcccccccc}
\toprule
\multirow{2}{*}{Method} & \multicolumn{4}{c}{\textbf{Area}} & \multicolumn{4}{c}{\textbf{Delay}} \\
\cmidrule(lr){2-5}\cmidrule(lr){6-9}
 & MAE$\downarrow$ & MAPE$\downarrow$ & $R^{2}$$\uparrow$ & RRSE$\downarrow$ & MAE$\downarrow$ & MAPE$\downarrow$ & $R^{2}$$\uparrow$ & RRSE$\downarrow$ \\
\midrule
CodeBERT           & 1.0011 & 0.17 & 0.3858 & 0.7837 & 0.7034 & 0.12 & 0.2539 & 0.8638 \\
CodeBERT-IP        & 0.9514 & 0.16 & 0.4646 & 0.7317 & 0.6576 & 0.11 & 0.3650 & 0.7969 \\
UniRTL (w/o graph) & 0.8818 & 0.15 & 0.5173 & 0.6948 & 0.6375 & 0.11 & 0.3839 & 0.7849 \\
\bottomrule
\end{tabular}
\end{table}

%% file: tabs/incomplete_pretrain_natural_language_code_search.tex
\begin{table}[t]
\centering
\caption{Performance comparison of UniRTL (w/o graph) and CodeBERT-based baselines on the natural language code search task, with F1 used as the main metric. CodeBERT-IP denotes the CodeBERT (Incomplete Pretrain).}
\label{tab:incomplete_pretrain_natural_language_code_search}
\begin{tabular}{lcccc}
\toprule
Model            & Design Format & Precision$\uparrow$ & Recall$\uparrow$ & F1$\uparrow$ (Main) \\ 
\midrule
CodeBERT         & code          & 0.565    & 0.625     & 0.585  \\
CodeBERT-IP      & code          & 0.600    & 0.658     & 0.618  \\
UniRTL (w/o graph) & code        & 0.630    & 0.683     & 0.644  \\
\bottomrule
\end{tabular}
\end{table}

%% file: tabs/incomplete_pretrain_functionality_equivalence_checking.tex
\begin{table}[t]
\centering
\caption{Performance comparison of UniRTL (w/o graph) and CodeBERT-based baselines on the functionality equivalence checking task, with average precision (AP) as the main metric. CodeBERT-IP denotes the CodeBERT (Incomplete Pretrain).}
\label{tab:incomplete_pretrain_functionality_equivalence_checking}
\setlength{\tabcolsep}{4pt}
\begin{tabular}{lcccccc}
\toprule
Model              & Design Format & AP$\uparrow$ (Main) & Accuracy$\uparrow$ & F1$\uparrow$ & Precision$\uparrow$ & Recall$\uparrow$ \\
\midrule
CodeBERT           & code          & 0.617 & 0.633 & 0.712 & 0.586 & 0.907 \\
CodeBERT-IP        & code          & 0.668 & 0.633 & 0.682 & 0.521 & 0.987 \\
UniRTL (w/o graph) & code          & 0.712 & 0.667 & 0.717 & 0.577 & 0.947 \\
\bottomrule
\end{tabular}
\vspace{-10pt}
\end{table}

%% file: tabs/alignment_ablation_performance_prediction.tex
\begin{table}[t]
\centering
\caption{Performance comparison of UniRTL variants with different code–graph alignment strategies on performance prediction tasks without incorporating netlist information. UniRTL (direct-combine) denotes a variant that simply concatenates the representations from the text–code-aligned encoder and the graph-aware tokenizer without explicit multimodal alignment pretraining.}
\label{tab:alignment_ablation_performance_prediction}
\setlength{\tabcolsep}{3.5pt}
\begin{tabular}{lcccccccc}
\toprule
\multirow{2}{*}{Method} & \multicolumn{4}{c}{\textbf{Area}} & \multicolumn{4}{c}{\textbf{Delay}} \\
\cmidrule(lr){2-5}\cmidrule(lr){6-9}
 & MAE$\downarrow$ & MAPE$\downarrow$ & $R^{2}$$\uparrow$ & RRSE$\downarrow$ 
 & MAE$\downarrow$ & MAPE$\downarrow$ & $R^{2}$$\uparrow$ & RRSE$\downarrow$ \\
\midrule
UniRTL (w/o graph)      & 0.8818 & 0.15 & 0.5173 & 0.6948 & 0.6375 & 0.11 & 0.3839 & 0.7849 \\
UniRTL (direct-combine) & 0.3629 & 0.06 & 0.7602 & 0.4897 & 0.3529 & 0.06 & 0.7624 & 0.4874 \\
UniRTL                  & 0.3510 & 0.06 & 0.7682 & 0.4815 & 0.3384 & 0.06 & 0.7832 & 0.4656 \\
\bottomrule
\end{tabular}
\end{table}

%% file: tabs/alignment_ablation_natural_language_code_search.tex
\begin{table}[t]
\centering
\caption{Performance comparison of UniRTL variants with different code–graph alignment strategies on the natural language code search task, with F1 used as the main metric. UniRTL (direct-combine) denotes a variant that simply concatenates the representations from the text–code-aligned encoder and the graph-aware tokenizer without explicit multimodal alignment pretraining.}
\label{tab:alignment_ablation_natural_language_code_search}
\begin{tabular}{lcccc}
\toprule
Model                 & Design Format & Precision$\uparrow$ & Recall$\uparrow$ & F1$\uparrow$ (Main) \\ 
\midrule
UniRTL (w/o graph)      & code            & 0.630 & 0.683 & 0.644 \\
UniRTL (direct-combine) & code \& graph   & 0.636 & 0.683 & 0.650 \\
UniRTL                  & code \& graph   & 0.650 & 0.692 & 0.662 \\
\bottomrule
\end{tabular}
\end{table}

%% file: tabs/alignment_ablation_functionality_equivalence_checking.tex
\begin{table}[t]
\centering
\caption{Performance comparison of UniRTL variants with different code–graph alignment strategies on the functionality equivalence checking task, with average precision (AP) as the main metric. UniRTL (direct-combine) denotes a variant that concatenates the representations from the text–code-aligned encoder and the graph-aware tokenizer without multimodal alignment pretraining.}
\label{tab:alignment_ablation_functionality_equivalence_checking}
\setlength{\tabcolsep}{3.7pt}
\begin{tabular}{lcccccc}
\toprule
Model                  & Design Format & AP$\uparrow$ (Main) & Accuracy$\uparrow$ & F1$\uparrow$ & Precision$\uparrow$ & Recall$\uparrow$ \\
\midrule
UniRTL (w/o graph)      & code          & 0.712 & 0.667 & 0.717 & 0.577 & 0.947 \\
UniRTL (direct-combine) & code \& graph & 0.737 & 0.687 & 0.723 & 0.595 & 0.920 \\
UniRTL                  & code \& graph & 0.745 & 0.747 & 0.753 & 0.734 & 0.773 \\
\bottomrule
\end{tabular}
\vspace{-10pt}
\end{table}

%% file: top.bib
@inproceedings{moravej2025graph,
  title={The graph's apprentice: teaching an LLM low-level knowledge for circuit quality estimation},
  author={Moravej, Reza and Bodhe, Saurabh and Zhang, Zhanguang and Ch{\'e}telat, Didier and Tsaras, Dimitrios and Zhang, Yingxue and Zhen, Hui-Ling and Hao, Jianye and Yuan, Mingxuan},
  booktitle={Proceedings of the Thirty-Fourth International Joint Conference on Artificial Intelligence},
  pages={9296--9304},
  year={2025}
}

@article{liu2025beyond,
  title={Beyond Tokens: Enhancing RTL Quality Estimation via Structural Graph Learning},
  author={Liu, Yi and Zhang, Hongji and Wang, Yiwen and Tsaras, Dimitris and Chen, Lei and Yuan, Mingxuan and Xu, Qiang},
  journal={arXiv preprint arXiv:2508.18730},
  year={2025}
}

@inproceedings{sengupta2022good,
  title={How good is your verilog rtl code? a quick answer from machine learning},
  author={Sengupta, Prianka and Tyagi, Aakash and Chen, Yiran and Hu, Jiang},
  booktitle={Proceedings of the 41st IEEE/ACM International Conference on Computer-Aided Design},
  pages={1--9},
  year={2022}
}

@inproceedings{liu2025deeprtl2,
  title={Deeprtl2: A versatile model for rtl-related tasks},
  author={Liu, Yi and Zhang, Hongji and Zhou, Yunhao and Shi, Zhengyuan and Xu, Changran and Xu, Qiang},
  booktitle={Findings of the Association for Computational Linguistics: ACL 2025},
  pages={6485--6500},
  year={2025}
}

@inproceedings{pei2024betterv,
  title={BetterV: controlled verilog generation with discriminative guidance},
  author={Pei, Zehua and Zhen, Hui-Ling and Yuan, Mingxuan and Huang, Yu and Yu, Bei},
  booktitle={Proceedings of the 41st International Conference on Machine Learning},
  pages={40145--40153},
  year={2024}
}

@inproceedings{liu2025craftrtl,
  title={Craftrtl: High-quality synthetic data generation for verilog code models with correct-by-construction non-textual representations and targeted code repair},
  author={Liu, Mingjie and Tsai, Yun-Da and Zhou, Wenfei and Ren, Haoxing},
  booktitle={International Conference on Learning Representations},
  volume={2025},
  pages={90377--90422},
  year={2025}
}

@inproceedings{liudeeprtl,
  title={DeepRTL: Bridging Verilog Understanding and Generation with a Unified Representation Model},
  author={Liu, Yi and Xu, Changran and Zhou, Yunhao and Li, Zeju and Xu, Qiang},
  booktitle={The Thirteenth International Conference on Learning Representations},
  year={2025}
}

@article{zhao2025codev,
  title={CodeV: Empowering LLMs with HDL Generation through Multi-Level Summarization},
  author={Zhao, Yang and Huang, Di and Li, Chongxiao and Jin, Pengwei and Song, Muxin and Xu, Yinan and Nan, Ziyuan and Gao, Mingju and Ma, Tianyun and Qi, Lei and others},
  journal={IEEE Transactions on Computer-Aided Design of Integrated Circuits and Systems},
  year={2025},
  publisher={IEEE}
}

@article{lewis2020retrieval,
  title={Retrieval-augmented generation for knowledge-intensive nlp tasks},
  author={Lewis, Patrick and Perez, Ethan and Piktus, Aleksandra and Petroni, Fabio and Karpukhin, Vladimir and Goyal, Naman and K{\"u}ttler, Heinrich and Lewis, Mike and Yih, Wen-tau and Rockt{\"a}schel, Tim and others},
  journal={Advances in neural information processing systems},
  volume={33},
  pages={9459--9474},
  year={2020}
}

@inproceedings{guo2021graphcodebert,
  title={GraphCodeBERT: Pre-training Code Representations with Data Flow},
  author={Guo, Daya and Ren, Shuo and Lu, Shuai and Feng, Zhangyin and Tang, Duyu and Liu, Shujie and Zhou, Long and Duan, Nan and Svyatkovskiy, Alexey and Fu, Shengyu and others},
  booktitle={The Ninth International Conference on Learning Representations},
  year={2021}
}

@inproceedings{fangcircuitfusion,
  title={CircuitFusion: Multimodal Circuit Representation Learning for Agile Chip Design},
  author={Fang, Wenji and Liu, Shang and Wang, Jing and Xie, Zhiyao},
  booktitle={The Thirteenth International Conference on Learning Representations},
  year={2025}
}

@article{vaswani2017attention,
  title={Attention is all you need},
  author={Vaswani, Ashish and Shazeer, Noam and Parmar, Niki and Uszkoreit, Jakob and Jones, Llion and Gomez, Aidan N and Kaiser, {\L}ukasz and Polosukhin, Illia},
  journal={Advances in neural information processing systems},
  volume={30},
  year={2017}
}

@inproceedings{velivckovic2018graph,
  title={Graph Attention Networks},
  author={Veli{\v{c}}kovi{\'c}, Petar and Cucurull, Guillem and Casanova, Arantxa and Romero, Adriana and Li{\`o}, Pietro and Bengio, Yoshua},
  booktitle={The Sixth International Conference on Learning Representations},
  year={2018}
}

@inproceedings{cui2024origen,
  title={Origen: Enhancing rtl code generation with code-to-code augmentation and self-reflection},
  author={Cui, Fan and Yin, Chenyang and Zhou, Kexing and Xiao, Youwei and Sun, Guangyu and Xu, Qiang and Guo, Qipeng and Liang, Yun and Zhang, Xingcheng and Song, Demin and others},
  booktitle={Proceedings of the 43rd IEEE/ACM International Conference on Computer-Aided Design},
  pages={1--9},
  year={2024}
}

@inproceedings{radford2021learning,
  title={Learning transferable visual models from natural language supervision},
  author={Radford, Alec and Kim, Jong Wook and Hallacy, Chris and Ramesh, Aditya and Goh, Gabriel and Agarwal, Sandhini and Sastry, Girish and Askell, Amanda and Mishkin, Pamela and Clark, Jack and others},
  booktitle={International conference on machine learning},
  pages={8748--8763},
  year={2021},
  organization={PmLR}
}

@article{li2021align,
  title={Align before fuse: Vision and language representation learning with momentum distillation},
  author={Li, Junnan and Selvaraju, Ramprasaath and Gotmare, Akhilesh and Joty, Shafiq and Xiong, Caiming and Hoi, Steven Chu Hong},
  journal={Advances in neural information processing systems},
  volume={34},
  pages={9694--9705},
  year={2021}
}

@inproceedings{li2022blip,
  title={Blip: Bootstrapping language-image pre-training for unified vision-language understanding and generation},
  author={Li, Junnan and Li, Dongxu and Xiong, Caiming and Hoi, Steven},
  booktitle={International conference on machine learning},
  pages={12888--12900},
  year={2022},
  organization={PMLR}
}

@inproceedings{li2023blip,
  title={Blip-2: Bootstrapping language-image pre-training with frozen image encoders and large language models},
  author={Li, Junnan and Li, Dongxu and Savarese, Silvio and Hoi, Steven},
  booktitle={International conference on machine learning},
  pages={19730--19742},
  year={2023},
  organization={PMLR}
}

@article{bao2022vlmo,
  title={Vlmo: Unified vision-language pre-training with mixture-of-modality-experts},
  author={Bao, Hangbo and Wang, Wenhui and Dong, Li and Liu, Qiang and Mohammed, Owais Khan and Aggarwal, Kriti and Som, Subhojit and Piao, Songhao and Wei, Furu},
  journal={Advances in neural information processing systems},
  volume={35},
  pages={32897--32912},
  year={2022}
}

@inproceedings{chuang2020speechbert,
  title={SpeechBERT: An Audio-and-Text Jointly Learned Language Model for End-to-End Spoken Question Answering},
  author={Chuang, Yung-Sung and Liu, Chi-Liang and Lee, Hung-yi and Lee, Lin-shan},
  booktitle={Proc. Interspeech 2020},
  pages={4168--4172},
  year={2020}
}

@inproceedings{tang2022unified,
  title={Unified Speech-Text Pre-training for Speech Translation and Recognition},
  author={Tang, Yun and Gong, Hongyu and Dong, Ning and Wang, Changhan and Hsu, Wei-Ning and Gu, Jiatao and Baevski, Alexei and Li, Xian and Mohamed, Abdelrahman and Auli, Michael and others},
  booktitle={Proceedings of the 60th Annual Meeting of the Association for Computational Linguistics (Volume 1: Long Papers)},
  pages={1488--1499},
  year={2022}
}

@inproceedings{yu2023speech,
  title={Speech-text pre-training for spoken dialog understanding with explicit cross-modal alignment},
  author={Yu, Tianshu and Gao, Haoyu and Lin, Ting-En and Yang, Min and Wu, Yuchuan and Ma, Wentao and Wang, Chao and Huang, Fei and Li, Yongbin},
  booktitle={Proceedings of the 61st Annual Meeting of the Association for Computational Linguistics (Volume 1: Long Papers)},
  pages={7900--7913},
  year={2023}
}

@inproceedings{jiangvlm2vec,
  title={VLM2Vec: Training Vision-Language Models for Massive Multimodal Embedding Tasks},
  author={Jiang, Ziyan and Meng, Rui and Yang, Xinyi and Yavuz, Semih and Zhou, Yingbo and Chen, Wenhu},
  booktitle={The Thirteenth International Conference on Learning Representations},
  year={2025}
}

@article{liu2024rtlcoder,
  title={Rtlcoder: Fully open-source and efficient llm-assisted rtl code generation technique},
  author={Liu, Shang and Fang, Wenji and Lu, Yao and Wang, Jing and Zhang, Qijun and Zhang, Hongce and Xie, Zhiyao},
  journal={IEEE Transactions on Computer-Aided Design of Integrated Circuits and Systems},
  volume={44},
  number={4},
  pages={1448--1461},
  year={2024},
  publisher={IEEE}
}

@inproceedings{zhang2024mg,
  title={Mg-verilog: Multi-grained dataset towards enhanced llm-assisted verilog generation},
  author={Zhang, Yongan and Yu, Zhongzhi and Fu, Yonggan and Wan, Cheng and Lin, Yingyan Celine},
  booktitle={2024 IEEE LLM Aided Design Workshop (LAD)},
  pages={1--5},
  year={2024},
  organization={IEEE}
}

@inproceedings{li2025deepcircuitx,
  title={Deepcircuitx: A comprehensive repository-level dataset for rtl code understanding, generation, and ppa analysis},
  author={Li, Zeju and Xu, Changran and Shi, Zhengyuan and Peng, Zedong and Liu, Yi and Zhou, Yunhao and Zhou, Lingfeng and Ma, Chengyu and Zhong, Jianyuan and Wang, Xi and others},
  booktitle={2025 IEEE International Conference on LLM-Aided Design (ICLAD)},
  pages={204--211},
  year={2025},
  organization={IEEE}
}

@inproceedings{wolf2013yosys,
  title={Yosys-a free verilog synthesis suite},
  author={Wolf, Clifford and Glaser, Johann and Kepler, Johannes},
  booktitle={Proceedings of the 21st Austrian Workshop on Microelectronics (Austrochip)},
  volume={97},
  pages={1--6},
  year={2013}
}

@inproceedings{chen2023incremental,
  title={Incremental verilog parser},
  author={Chen, Xiangli and Meng, Yuehua and Chen, Gang},
  booktitle={2023 International Symposium of Electronics Design Automation (ISEDA)},
  pages={236--240},
  year={2023},
  organization={IEEE}
}

@inproceedings{feng2020codebert,
  title={CodeBERT: A Pre-Trained Model for Programming and Natural Languages},
  author={Feng, Zhangyin and Guo, Daya and Tang, Duyu and Duan, Nan and Feng, Xiaocheng and Gong, Ming and Shou, Linjun and Qin, Bing and Liu, Ting and Jiang, Daxin and others},
  booktitle={Findings of the Association for Computational Linguistics: EMNLP 2020},
  pages={1536--1547},
  year={2020}
}

@article{husain2019codesearchnet,
  title={Codesearchnet challenge: Evaluating the state of semantic code search},
  author={Husain, Hamel and Wu, Ho-Hsiang and Gazit, Tiferet and Allamanis, Miltiadis and Brockschmidt, Marc},
  journal={arXiv preprint arXiv:1909.09436},
  year={2019}
}

@inproceedings{devlin2019bert,
  title={Bert: Pre-training of deep bidirectional transformers for language understanding},
  author={Devlin, Jacob and Chang, Ming-Wei and Lee, Kenton and Toutanova, Kristina},
  booktitle={Proceedings of the 2019 conference of the North American chapter of the association for computational linguistics: human language technologies, volume 1 (long and short papers)},
  pages={4171--4186},
  year={2019}
}

@inproceedings{xupowerful,
  title={How Powerful are Graph Neural Networks?},
  author={Xu, Keyulu and Hu, Weihua and Leskovec, Jure and Jegelka, Stefanie},
  booktitle={The Seventh International Conference on Learning Representations},
  year={2019}
}

@article{mackiewicz1993principal,
  title={Principal components analysis (PCA)},
  author={Ma{\'c}kiewicz, Andrzej and Ratajczak, Waldemar},
  journal={Computers \& Geosciences},
  volume={19},
  number={3},
  pages={303--342},
  year={1993},
  publisher={Elsevier}
}

@article{rampavsek2022recipe,
  title={Recipe for a general, powerful, scalable graph transformer},
  author={Ramp{\'a}{\v{s}}ek, Ladislav and Galkin, Michael and Dwivedi, Vijay Prakash and Luu, Anh Tuan and Wolf, Guy and Beaini, Dominique},
  journal={Advances in Neural Information Processing Systems},
  volume={35},
  pages={14501--14515},
  year={2022}
}

@book{chung1997spectral,
  title={Spectral graph theory},
  author={Chung, Fan RK},
  volume={92},
  year={1997},
  publisher={American Mathematical Soc.}
}

@inproceedings{brayton2010abc,
  title={ABC: An academic industrial-strength verification tool},
  author={Brayton, Robert and Mishchenko, Alan},
  booktitle={International Conference on Computer Aided Verification},
  pages={24--40},
  year={2010},
  organization={Springer}
}

@inproceedings{edwards2020google,
  title={Google/SkyWater and the Promise of the Open PDK},
  author={Edwards, R Timothy},
  booktitle={Workshop on Open-Source EDA Technology},
  volume={124},
  pages={132},
  year={2020}
}

@article{guo2024deepseek,
  title={DeepSeek-Coder: When the Large Language Model Meets Programming--The Rise of Code Intelligence},
  author={Guo, Daya and Zhu, Qihao and Yang, Dejian and Xie, Zhenda and Dong, Kai and Zhang, Wentao and Chen, Guanting and Bi, Xiao and Wu, Yu and Li, YK and others},
  journal={arXiv preprint arXiv:2401.14196},
  year={2024}
}

@article{roziere2023code,
  title={Code llama: Open foundation models for code},
  author={Roziere, Baptiste and Gehring, Jonas and Gloeckle, Fabian and Sootla, Sten and Gat, Itai and Tan, Xiaoqing Ellen and Adi, Yossi and Liu, Jingyu and Sauvestre, Romain and Remez, Tal and others},
  journal={arXiv preprint arXiv:2308.12950},
  year={2023}
}

@article{bai2023qwen,
  title={Qwen technical report},
  author={Bai, Jinze and Bai, Shuai and Chu, Yunfei and Cui, Zeyu and Dang, Kai and Deng, Xiaodong and Fan, Yang and Ge, Wenbin and Han, Yu and Huang, Fei and others},
  journal={arXiv preprint arXiv:2309.16609},
  year={2023}
}

@article{grattafiori2024llama,
  title={The llama 3 herd of models},
  author={Grattafiori, Aaron and Dubey, Abhimanyu and Jauhri, Abhinav and Pandey, Abhinav and Kadian, Abhishek and Al-Dahle, Ahmad and Letman, Aiesha and Mathur, Akhil and Schelten, Alan and Vaughan, Alex and others},
  journal={arXiv preprint arXiv:2407.21783},
  year={2024}
}

@article{neelakantan2022text,
  title={Text and code embeddings by contrastive pre-training},
  author={Neelakantan, Arvind and Xu, Tao and Puri, Raul and Radford, Alec and Han, Jesse Michael and Tworek, Jerry and Yuan, Qiming and Tezak, Nikolas and Kim, Jong Wook and Hallacy, Chris and others},
  journal={arXiv preprint arXiv:2201.10005},
  year={2022}
}

@inproceedings{lee2025nv,
  title={Nv-embed: Improved techniques for training llms as generalist embedding models},
  author={Lee, Chankyu and Roy, Rajarshi and Xu, Mengyao and Raiman, Jonathan and Shoeybi, Mohammad and Catanzaro, Bryan and Ping, Wei},
  booktitle={International Conference on Learning Representations},
  volume={2025},
  pages={79310--79333},
  year={2025}
}

@inproceedings{muennighoff2025generative,
  title={Generative representational instruction tuning},
  author={Muennighoff, Niklas and Su, Hongjin and Wang, Liang and Yang, Nan and Wei, Furu and Yu, Tao and Singh, Amanpreet and Kiela, Douwe},
  booktitle={International Conference on Learning Representations},
  volume={2025},
  pages={45544--45613},
  year={2025}
}

@inproceedings{muennighoff2023mteb,
  title={Mteb: Massive text embedding benchmark},
  author={Muennighoff, Niklas and Tazi, Nouamane and Magne, Lo{\"\i}c and Reimers, Nils},
  booktitle={Proceedings of the 17th Conference of the European Chapter of the Association for Computational Linguistics},
  pages={2014--2037},
  year={2023}
}

@article{oord2018representation,
  title={Representation learning with contrastive predictive coding},
  author={Oord, Aaron van den and Li, Yazhe and Vinyals, Oriol},
  journal={arXiv preprint arXiv:1807.03748},
  year={2018}
}

@inproceedings{loshchilovdecoupled,
  title={Decoupled Weight Decay Regularization},
  author={Loshchilov, Ilya and Hutter, Frank},
  booktitle={The Seventh International Conference on Learning Representations},
  year={2019}
}

@article{saleh2022statistical,
  title={Statistical properties of the log-cosh loss function used in machine learning},
  author={Saleh, Resve A and Saleh, AK},
  journal={arXiv preprint arXiv:2208.04564},
  year={2022}
}

@inproceedings{kingma2014adam,
  title={Adam: A method for stochastic optimization},
  author={Kingma, Diederik P and Ba, Jimmy},
  booktitle={The Third International Conference on Learning Representations},
  year={2015}
}

@inproceedings{cui2019class,
  title={Class-balanced loss based on effective number of samples},
  author={Cui, Yin and Jia, Menglin and Lin, Tsung-Yi and Song, Yang and Belongie, Serge},
  booktitle={Proceedings of the IEEE/CVF conference on computer vision and pattern recognition},
  pages={9268--9277},
  year={2019}
}
